\pdfoutput=1

\documentclass[10pt,twocolumn,letterpaper]{article}

\usepackage[pagenumbers]{cvpr} 

\makeatletter
\renewcommand{\paragraph}{
  \@startsection{paragraph}{4}
  {\z@}{1.0ex \@plus 0ex \@minus .3ex}{-1em}
  {\normalfont\normalsize\bfseries}
}
\makeatother
\usepackage{graphicx}
\usepackage{amsmath}
\usepackage{amssymb}
\usepackage{booktabs}
\usepackage{multirow}
\usepackage{bbm}
\usepackage{bbding}
\usepackage[utf8]{inputenc}
\usepackage[misc]{
    ifsym}

\usepackage[accsupp]{axessibility}  

\usepackage[pagebackref,breaklinks,colorlinks]{hyperref}

\usepackage[capitalize]{cleveref}
\crefname{section}{Sec.}{Secs.}
\Crefname{section}{Section}{Sections}
\Crefname{table}{Table}{Tables}
\crefname{table}{Tab.}{Tabs.}

\begin{document}

\title{GAPartNet: Cross-Category Domain-Generalizable Object Perception\\ and Manipulation via Generalizable and Actionable Parts}

\author{
Haoran Geng\footnotemark[1] \textsuperscript{ 1,2,3} \quad 
Helin Xu\footnotemark[1] \textsuperscript{ 4} \quad
Chengyang Zhao\footnotemark[1] \textsuperscript{ 1} \\
Chao Xu \textsuperscript{5} \quad 
Li Yi \textsuperscript{4} \quad 
Siyuan Huang \textsuperscript{3} \quad
He Wang\footnotemark[2] \textsuperscript{ 1,2} \\
\textsuperscript{1}CFCS, Peking University \quad
\textsuperscript{2}School of EECS, Peking University \\
\textsuperscript{3}Beijing Institute for General Artificial Intelligence   \\
\textsuperscript{4}Tsinghua University \quad
\textsuperscript{5}University of California, Los Angeles \\
\href{https://pku-epic.github.io/GAPartNet}{https://pku-epic.github.io/GAPartNet}
}

\twocolumn[{%
\renewcommand\twocolumn[1][]{#1}%
\maketitle
\begin{center}
    \centering
    \captionsetup{type=figure}
    \includegraphics[width=0.85\linewidth]{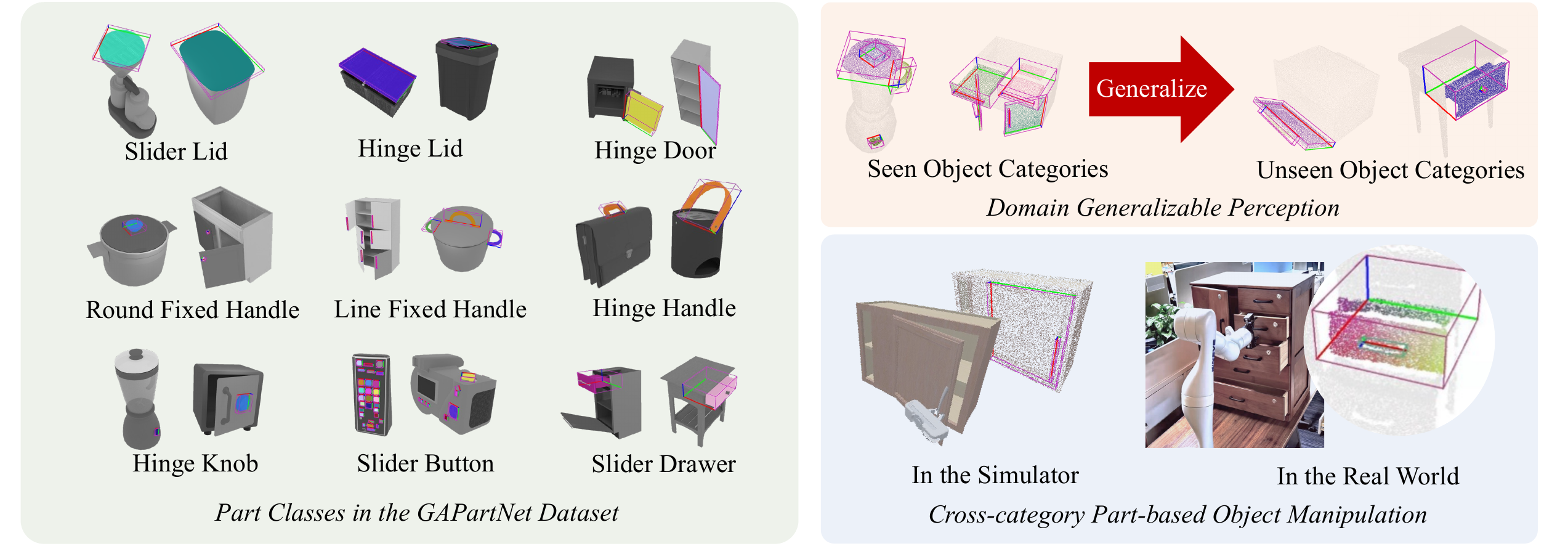}
    \captionof{figure}{\textbf{Overview.} We propose to learn generalizable object perception and manipulation skills via \textbf{G}eneralizable and \textbf{A}ctionable \textbf{Parts}, and present \textbf{GAPartNet}, a large-scale interactive dataset with rich part annotations. We propose a domain generalization method for cross-category part segmentation and pose estimation. Our GAPart definition boosts cross-category object manipulation and can transfer to real.}
    \label{fig:teaser}
\end{center}
}]

\renewcommand{\thefootnote}{\fnsymbol{footnote}}
\footnotetext[1]{Equal contribution with the order determined by rolling dice.}
\footnotetext[2]{Corresponding author: \href{mailto:hewang@pku.edu.cn}{hewang@pku.edu.cn}.}

\begin{abstract}
For years, researchers have been devoted to generalizable object perception and manipulation, where cross-category generalizability is highly desired yet underexplored.
In this work, we propose to learn such cross-category skills via \textbf{G}eneralizable and \textbf{A}ctionable \textbf{Parts} (\textbf{GAParts}). 
By identifying and defining 9 GAPart classes (lids, handles, etc.) in 27 object categories, we construct a large-scale part-centric interactive dataset, GAPartNet,
where we provide rich, part-level annotations (semantics, poses) for 8,489 part instances on 1,166 objects.
Based on GAPartNet, we investigate three cross-category tasks: part segmentation, part pose estimation, and part-based object manipulation.
Given the significant domain gaps between seen and unseen object categories, we
propose a robust 3D segmentation method from the perspective of domain generalization by integrating adversarial learning techniques. Our method outperforms all existing methods by a large margin, no matter on seen or unseen categories. Furthermore, with part segmentation and pose estimation results, we leverage the GAPart pose definition to design part-based manipulation heuristics that can generalize well to unseen object categories in both the simulator and the real world.
Our dataset, code, and demos are available on our project page.

\end{abstract}

\section{Introduction \label{sec:intro}}
Generalizable object perception and manipulation are at the core of building intelligent and multi-functional robots. Recent efforts on generalizing the vision have been devoted to category-level object perception that deals with perceiving novel object instances from known object categories, including object detectors from RGB images~\cite{ren2015faster,he2017mask,kirillov2020pointrend}, point clouds~\cite{jiang2020pointgroup,chen2021hierarchical}, and category-level pose estimation works on rigid~\cite{wang2019normalized,Chen_2020_CVPR} and articulated objects~\cite{li2020category,weng2021captra}. On the front of generalizable manipulation, complex tasks that involve interacting with articulated objects have also been proposed in a category-level fashion, as in the recent challenge on learning category-level manipulation skills \cite{mu2021maniskill}. Additionally, to boost robot perception and manipulation with indoor objects, researchers have already proposed several datasets \cite{yi2016scalable,mo2019partnet,xiang2020sapien,yu2019partnet,wang2019shape2motion} with part segmentation and motion annotations, and have devoted work to part segmentation \cite{mo2019partnet,yu2019partnet} and articulation estimation~\cite{li2020category}.

However, these works all approach the object perception and manipulation problems in an intra-category manner, while humans can well perceive and interact with instances from unseen object categories based on prior knowledge of functional parts such as buttons, handles, lids, \etc. In fact, parts from the same classes have fewer variations in their shapes and the ways that we manipulate them, compared to objects from the same categories. 
We thus argue that part classes are more elementary and fundamental compared to object categories, and generalizable visual perception and manipulation tasks should be conducted at part-level.

Then, what defines a part class? Although there is no single answer, we propose to identify part classes that are generalizable in both recognition and manipulation. After careful thoughts and expert designs, we propose the concept of \textit{\textbf{G}eneralizable and \textbf{A}ctionable \textbf{Part} (\textbf{GAPart})} classes. Parts from the same GAPart class share similar shapes which allow generalizable visual recognition; parts from the same GAPart class also have aligned actionability and can be interacted with in a similar way, which ensures minimal human effort when designing interaction guidance to achieve generalizable and robust manipulation policies.

Along with the GAPart definition, we present GAPartNet, a large-scale interactive part-centric dataset where we gather 1,166 articulated objects from the PartNet-Mobility dataset \cite{xiang2020sapien} and the AKB-48 dataset \cite{liu2022akb}. We put in great effort in identifying and annotating semantic labels to 8,489 GAPart instances. Moreover, we systematically align and annotate the GAPart poses, which we believe serve as the bridge between visual perception and manipulation. Our class-level GAPart pose definition highly couples the part poses with how we want to interact with the parts. We show that this is highly desirable -- once the part poses are known, we can easily manipulate the parts using simple heuristics.

Based on the proposed dataset, we further explore three cross-category tasks based on GAParts: part segmentation, part pose estimation, and part-based object manipulation, where we aim at recognizing and interacting with the parts from novel objects in both known categories and, moreover, unseen object categories.  In this work, we propose to use learning-based methods to deal with perception tasks, after which, based on the GAPart definition, we devise simple heuristics to achieve cross-category object manipulation. 

However, different object categories may contain different kinds of GAParts and provide different contexts for the parts. 
Each object category thus forms a unique domain for perceiving and manipulating GAParts.
Therefore, all three tasks demand domain-generalizable methods that can work on unseen object categories without seeing them during training, which is very challenging for existing vision and robotic algorithms. We thus consult the generalization literature \cite{ganin2016domain,ganin2015unsupervised,li2018domain} and propose to learn domain-invariant representation, which is often achieved by domain adversarial learning with a domain classifier. During training, the classifier tries to distinguish the domains while the feature extractor tries to fool the classifier, which encourages domain-invariant feature learning.
However, it is highly non-trivial to adopt adversarial learning in our domain-invariant feature learning, due to the following challenges.
1) \textit{Handling huge variations in part contexts across different domains.} 
The context of a GAPart class can vary significantly across different object categories.
For example, in training data, round handles usually sit on the top of lids for the CoffeeMachine category, whereas for the test category Table, round handles often stand to the front face of the drawers.
To robustly segment GAParts in objects from unseen categories, we need the part features to be context-invariant.
2) \textit{Handling huge variations in part sizes.} Parts from different GAPart classes may be in different sizes, \eg, a button is usually much smaller than a door. Given that the input is a point cloud, the variations in part sizes will result in huge variations in the number of points across different GAParts, which makes feature learning very challenging.
3) \textit{Handling the imbalanced part distribution and part-object relations.} Object parts in the real world distribute naturally unevenly and a particular part class may appear with different frequencies throughout various object categories. For example, there can be more buttons than doors on a washing machine while the opposite is true in the case of a storage furniture. This imbalanced distribution also adds difficulties to the learning of domain-invariant features.

Accordingly, we integrate several important techniques from domain adversarial learning. To improve context invariance, we propose a part-oriented feature query technique that mainly focuses on foreground parts and ignores the background. To handle diverse part sizes, we propose a multi-resolution technique. Finally, we employ the focal loss to handle the distribution imbalance. Our method significantly outperforms previous 3D instance segmentation methods and achieves 76.5\% AP50 on seen object categories and 37.2\% AP50 on unseen categories.

To summarize, our main contributions are as follows:

1. We provide the concept of \textit{GAPart} 
and present a large-scale interactive dataset, \textit{GAPartNet}, with rich part semantics and pose annotations that facilitates generalizable part perception and part-based object manipulation.

2. We propose a first-ever pipeline for domain-generalizable 3D part segmentation and pose estimation via learning domain-invariant features, which
significantly outperforms the baselines.

3. We provide a new solution to generalizable object manipulation by leveraging the concept of GAParts. Thanks to innate generalizability and actionability, minimal human effort is needed when designing interaction guidance to achieve generalizable and robust manipulation policies.

\section{Related Work\label{sec:relwork}}

\paragraph{Part Instance Segmentation from Point Cloud Observations.}
Large-scale datasets of 3D shapes are fundamental to 3D part segmentation works, \eg, ShapeNet (2$\sim$5 parts per object) \cite{chang2015shapenet,yi2016scalable} and PartNet (15 parts per object on average) \cite{mo2019partnet}. Based on such datasets, much progress has been made on unified architectures for point cloud learning \cite{qi2017pointnet,qi2017pointnet++,li2018pointcnn,wang2019dynamic}, specialized supervised segmentation networks \cite{wang2018sgpn,yi2019gspn}, shape abstraction and part discovery \cite{mo2019structurenet,paschalidou2021neural,yang2021unsupervised,xu2022partafford}, \textit{etc}. However, these works all approach object perception in an intra-category manner. We instead tackle 3D part instance segmentation in a cross-category way, based on our newly proposed GAPartNet dataset.

\paragraph{Domain Generalization.} 
To tackle the out-of-distribution problems, domain generalization methods try to learn from multiple source domains to generalize to the unseen domains, which can be divided into the following three categories\cite{DG_survey}: 1) data manipulation methods (\eg, data augmentation\cite{zhou2021domain}, data generation\cite{zhou2020learning,rahman2019multi}); 2) learning strategy design (\eg, ensemble learning\cite{xu2014exploiting,zhou2021domain}, meta learning\cite{li2018learning,li2019episodic},  automated machine learning\cite{cortes2017adanet}); 3) domain-invariant representation learning (\eg, explicit feature alignment\cite{peng2019moment,zhou2020domain}, domain adversarial learning\cite{li2018domain,li2018deep,muandet2013domain,ganin2016domain,ganin2015unsupervised}). 
However, works on domain generalization mainly focus on 2D tasks (\eg, image classification), whose techniques are not suitable to be directly used in our 3D multi-stage part segmentation and pose estimation tasks. \cite{liu2022autogpart,luo2020learning} try to discover parts in a category-agnostic manner, but their task settings are also different from ours. In our tasks, we need to tackle irregular point cloud representation and take the multi-stage, multi-part setting into account.

\paragraph{Category-level Object Pose Estimation.}
Pose estimation has been studied at instance-level as well as category-level. Instance-level object pose estimation works~\cite{xiang2018posecnn,li2018deepim,sundermeyer2018implicit,peng2019pvnet,wang2019densefusion,labbe2020cosypose,He_2021_CVPR} assume known CAD models and thus have their limitations. Other works, on the other hand, deal with 3D bounding boxes prediction and 6D pose estimation at category-level, including single-frame pose estimation such as NOCS \cite{wang2019normalized}, FS-Net \cite{Chen_2021_CVPR}, CASS \cite{Chen_2020_CVPR,wang2021category}, and category-level tracking such as 9-PACK\cite{wang20206}, CAPTRA\cite{weng2021captra}. Wang \textit{et al.}\cite{wang2019normalized} innovates Normalized Object Coordinate Space (NOCS), a unified coordinate space where objects from the same category are normalized, canonicalized and share an identical orientation.
CASS \cite{Chen_2020_CVPR} learns a canonical latent shape space for certain object categories, while \cite{Tian_2020_ECCV} leverages category shape priors and models shape deformations to handle intra-class shape variations. FS-Net \cite{Chen_2021_CVPR} designs a fast shape-based network that extracts efficient category-level pose features. \cite{wang2021category} uses a cascaded relation network to relate 2D, 3D, shape priors, and proposes a recurrent reconstruction network to make iterative improvements.

\paragraph{Generalizable Object Manipulation.}
On the front of object manipulation, Mu \textit{et al.} proposes \cite{mu2021maniskill} a challenge to learn generalizable manipulation skills for articulated objects from known categories. Although some previous methods\cite{mo2021where2act, geng2022end, gong2022arnold} have certain generalizability, robotic manipulation in a novel environment still calls for the ability to handle novel object categories. Although, for simple rigid objects, there is existing literature on robust and object-agnostic object grasping \cite{fang2020graspnet,jiang2021synergies,breyer2020volumetric} and planar pushing \cite{li2018push,yu2016more} algorithms,
while very few works have been devoted to interacting with articulated objects that contain movable parts. 
Recently, Mo \textit{et al.}\cite{mo2021where2act} and Wu \textit{et al.}   \cite{wu2022vatmart} tackle this problem by leveraging low-level generalizability.
The most related work to us is Gadre \textit{et al.} \cite{gadre2021act} which proposes an interactive perception pipeline learning to touch, watch, then segment the object into movable parts.
However, this work does not consider the consistent geometry and actionability patterns behind parts from the same class and can only deal with simple objects with up to three parts on the table surfaces, \eg, scissors and eyeglasses.

\begin{figure}[!ht]
    \centering
    \includegraphics[width=1\linewidth]{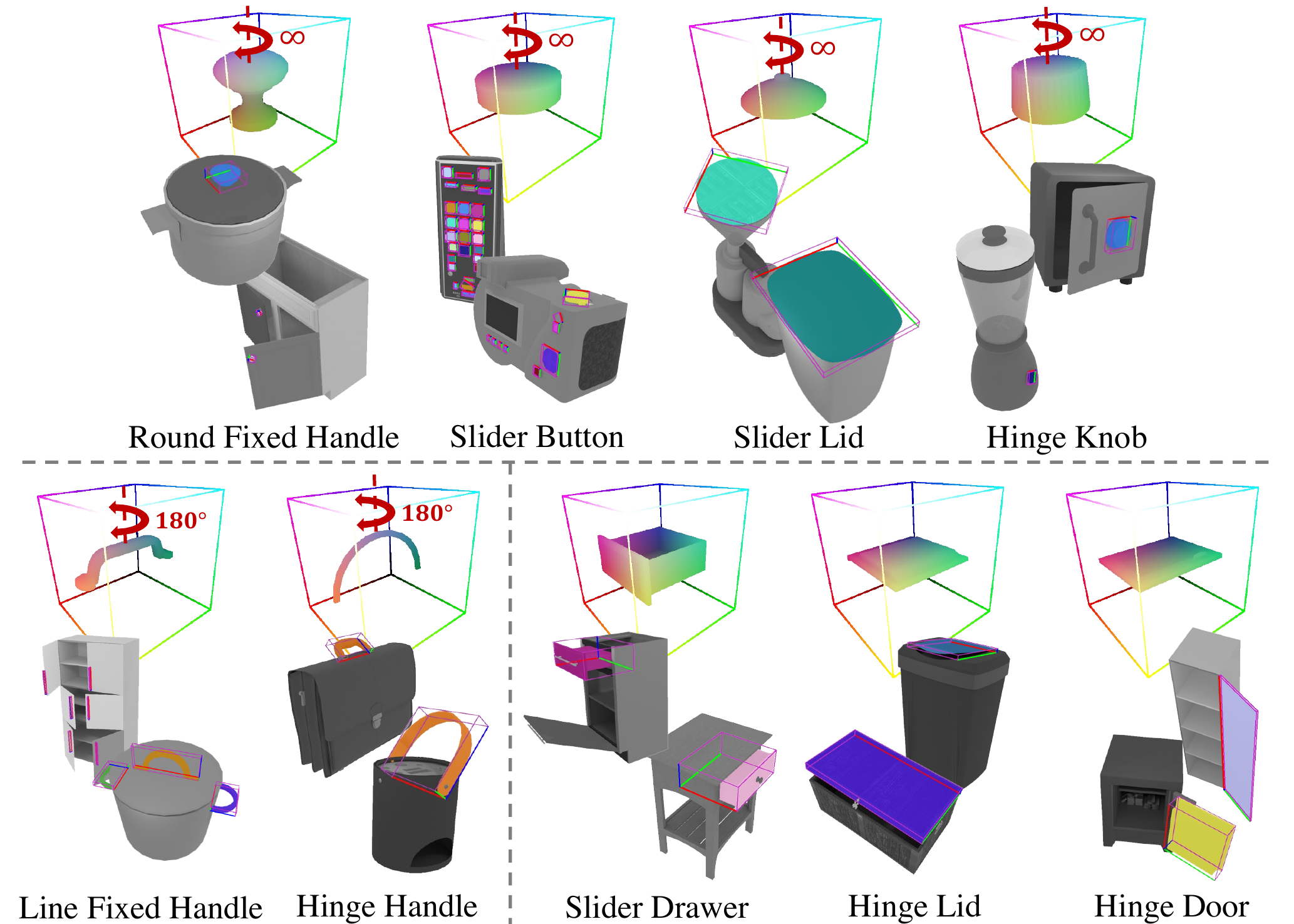}
    \caption{\textbf{GAPart Classes.} Here we highlight the parts from 9 GAPart classes along with their normalized part coordinate spaces. On the {top}, we show the four GAPart classes that have continuous rotation symmetry along the $z$ axis, denoted with the red-dashed line and the $\infty$ remark; the {bottom-left} shows the two GAPart classes that have ${180}^{\circ}$ mirror symmetry along the $z$ axis; and the {bottom-right} shows the rest three asymmetric GAPart classes.}
    \label{fig:partdefinition}
\end{figure}

\begin{figure*}[!ht]
    \centering
    \includegraphics[width=1\linewidth]{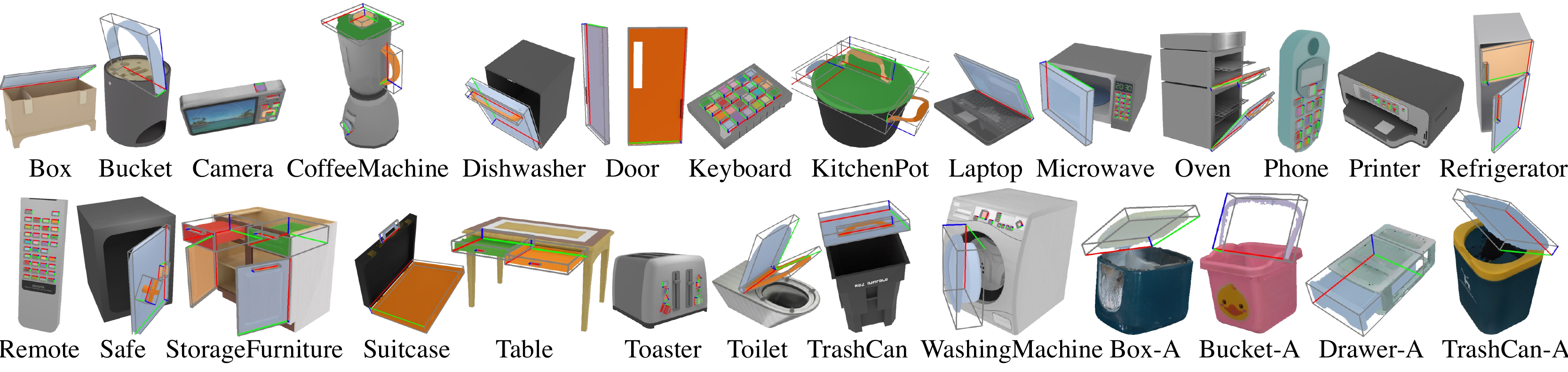}
    \caption{\textbf{GAPartNet Objects.} Objects collected from AKB-48 \cite{liu2022akb} end with '-A', while the others are from PartNet-Mobility \cite{xiang2020sapien}.}
    \label{fig:dataset overview}
\end{figure*}

\begin{table*}
\centering
\setlength\tabcolsep{2pt}
\resizebox{.99\textwidth}{!}{

\begin{tabular}{c|c|ccccccccccccccccccccccccccc}
\hline
& \textbf{All}  & \textbf{Bo} & \textbf{Bu} & \textbf{Ca} & \textbf{Co} & \textbf{Di} & \textbf{Do} & \textbf{Ke} & \textbf{Ki} & \textbf{La} & \textbf{Mi} & \textbf{Ov} & \textbf{Ph} & \textbf{Pr} & \textbf{Ref} & \textbf{Rem} & \textbf{Sa} & \textbf{St} & \textbf{Su} & \textbf{Ta} & \textbf{Toa} & \textbf{Toi} & \textbf{Tr} & \textbf{Wa} & \textbf{Bo-A} & \textbf{Bu-A} & \textbf{Dr-A} & \textbf{Tr-A}  \\
\hline
\textbf{Object}  & 1,166   & 25    & 31    & 32    & 41    & 41    & 14    & 31    & 20    & 48    & 16    & 23    & 15    & 28    & 37    & 49    & 29    & 324   & 10    & 77    & 19    & 66    & 52    & 17    & 40    & 37    & 22    & 22    \\
\hline
\textbf{Ln.F.Hl.} & 922   & 2     &  \textendash     &  \textendash     & 10    & 28    & 2     &  \textendash    & 40    &  \textendash     & 5     & 29    &  \textendash    &    \textendash   & 43    &   \textendash    & 1     & 667   &   \textendash    & 60    & \textendash      &   \textendash    & 35    &  \textendash     &   \textendash    &  \textendash     &   \textendash    &  \textendash     \\
\textbf{Rd.F.Hl.} & 151   &   \textendash    &    \textendash   &    \textendash   & 8     &    \textendash   & 9     &   \textendash    & 14    &   \textendash    &    \textendash   &   \textendash    &  \textendash     &   \textendash    &   \textendash    &   \textendash    &    \textendash   & 54    &   \textendash    & 65    &   \textendash    &   \textendash    & 1     &   \textendash    &   \textendash    &    \textendash   &   \textendash    &   \textendash    \\
\textbf{Hg.Hl.} & 78    &   \textendash    & 31    &    \textendash   &    \textendash   &   \textendash    &   \textendash    &   \textendash    &  \textendash     &   \textendash    &   \textendash    &   \textendash    &   \textendash    &   \textendash    &   \textendash    &  \textendash     &   \textendash    &   \textendash    & 10    &   \textendash    &   \textendash    &   \textendash    &   \textendash    &   \textendash    &   \textendash    & 37    &   \textendash    &    \textendash   \\
\textbf{Hg.Ld.}  & 260   & 49    &   \textendash    &   \textendash    & 1     &   \textendash    &    \textendash   &   \textendash    &   \textendash    & 48    &   \textendash    & 1     &   \textendash    &   \textendash    &   \textendash    &   \textendash    &   \textendash    & 1     & 7     &   \textendash    &   \textendash    & 55    & 31    & 5     & 40    &   \textendash    &    \textendash   & 22    \\
\textbf{Sd.Ld.}  & 89    &  \textendash     &   \textendash    & 1     & 19    &   \textendash    &   \textendash    &   \textendash    & 20    &   \textendash    &    \textendash   &   \textendash    &   \textendash    &   \textendash    &    \textendash   &   \textendash    &   \textendash    &   \textendash    &   \textendash    &   \textendash    &   \textendash    & 44    & 5     &   \textendash    &   \textendash    &   \textendash    &   \textendash    &   \textendash    \\
\textbf{Sd.Bn.} & 5,526  &   \textendash    &   \textendash    & 208   & 140   & 5     &   \textendash    & 2,934   & 1     &    \textendash   & 39    & 17    & 227   & 311   &   \textendash    & 1,433   & 89    &    \textendash   & 2     &   \textendash    & 24    & 15    &    \textendash   & 81    &   \textendash    &   \textendash    &   \textendash    &   \textendash    \\
\textbf{Sd.Dw.} & 546   & 1     &   \textendash    &   \textendash    &   \textendash    & 1     &    \textendash   &   \textendash    &   \textendash    &   \textendash    &    \textendash   &  \textendash     &  \textendash     & 6     &   \textendash    &   \textendash    &   \textendash    & 333   &   \textendash    & 161   &   \textendash    &   \textendash    & 7     &   \textendash    &   \textendash    &  \textendash     & 37    &  \textendash     \\
\textbf{Hg.Dr.} & 678   &  \textendash     &   \textendash    &   \textendash    &  \textendash     & 41    & 18    &   \textendash    &  \textendash     &   \textendash    & 16    & 28    &   \textendash    &    \textendash   & 60    &   \textendash    & 29    & 433   &    \textendash   & 24    &   \textendash    &    \textendash   & 15    & 14    &    \textendash   &    \textendash   &   \textendash    &   \textendash    \\
\textbf{Hg.Kb.} & 239   &   \textendash    &    \textendash   & 11    & 47    & 2     &   \textendash    &   \textendash    &   \textendash    &  \textendash     & 8     & 77    &   \textendash    & 1     &   \textendash    & 1     & 37    &  \textendash     &   \textendash    &   \textendash    & 28    &  \textendash     &   \textendash    & 27    &   \textendash    &    \textendash   &   \textendash    &    \textendash   \\
\hline
\end{tabular}
}
\caption{\textbf{GAPartNet Statistics.} We show how the GAPart instances distribute across all object categories, where object categories titles are in the same order as in \cref{fig:dataset overview}, \eg, \textbf{Bo} = Box, \textbf{Bu} = Bucket, \etc. The first row \textbf{Object} shows the object number for each object category. The following rows, \ie, line fixed handle, round fixed handle, hinge handle, hinge lid, slider lid, slider button, slider drawer, hinge door, and hinge knob, show the number of GAPart instances in each object category.}
\label{table:datasetstatistics}
\end{table*}

\section{GAPart Definition and GAPartNet Dataset}
\label{sec:dataset_def}
\subsection{GAPart Definition}



Different from previous works, we give a rigorous definition to the GAPart classes, which not only are Generalizable to visual recognition but also share similar Actionability, corresponding to the G and A in GAPartNet. Our main purpose of such a definition is to bridge the perception and manipulation, to allow joint learning of both vision and interaction. Accordingly, we propose two principles to follow: \textbf{firstly, geometric similarity within part classes,} and \textbf{secondly, actionability alignment within part classes.}

\paragraph{GAPart Semantics.} Based on such principles, we identify 9 common GAPart classes across 27 object categories: \textit{line fixed handle, round fixed handle, hinge handle, hinge lid, slider lid, slider button, slider drawer, hinge door, hinge knob}. 
Note that based on different actionability, handles are split into \textit{fixed} handles and \textit{hinge} handles, while lids are split into \textit{hinge} lids and \textit{slider} lids. We further identify \textit{line} fixed handles and \textit{round} fixed handles according to their difference in geometry.

\paragraph{GAPart Poses.} Following previous works\cite{wang2019normalized,li2020category}, we define the canonicalized part position and orientation in Normalized Part Coordinate Space (NPCS) for each GAPart class. We illustrate our pose definition in \cref{fig:partdefinition}. Note that some of the GAPart classes have innate symmetry, which should be taken care of when dealing with their poses.

Based on the rigorous and manipulation-oriented definition, simple heuristics can be designed to achieve generalizable part-based manipulation across different object categories, once we know the part classes and the part poses.

\subsection{GAPartNet Dataset}
Following the GAPart definition, we construct a large-scale part-centric interactive dataset, GAPartNet, with rich, part-level annotations for both perception and interaction tasks. Our 3D object shapes come from two existing datasets, PartNet-Mobility\cite{xiang2020sapien} and AKB-48\cite{liu2022akb}, which are cleaned and provided with new uniform annotations based on our GAPart definition. The final GAPartNet has 9 GAPart classes, providing semantic labels and pose annotations for 8,489 GAPart instances on 1,166 objects from 27 object categories. On average, each object has 7.3 functional parts. Each GAPart class can be seen on objects from more than 3 object categories, and each GAPart class is found in 8.8 object categories on average, which lays the foundation for our benchmark on generalizable parts.

\cref{table:datasetstatistics} and \cref{fig:dataset overview} show the statistics and selected examples of GAPartNet. 

\subsection{Data Annotation}

We direct systemic works to guarantee cross-category generalizable part semantics and pose annotations. We follow the steps below to clean and annotate our data: 1) Fixing imperfect meshes and re-merging the meshes into new parts. The average fixing time per object is 15 minutes, while the average re-merging time per part is 5 minutes. Over 100 object instances are fixed and over 1,000 GAPart instances are newly merged. 2) Annotating cross-category semantic labels. 3) Aligning and Annotating poses. We spend more than 200 hours building a whole pipeline as well as several manually designed rules to align and annotate the poses of all GAParts. 

More dataset visualizations and annotation details can be found in the appendix.

\begin{figure*}[!t]
\centering
\includegraphics[width=\linewidth]{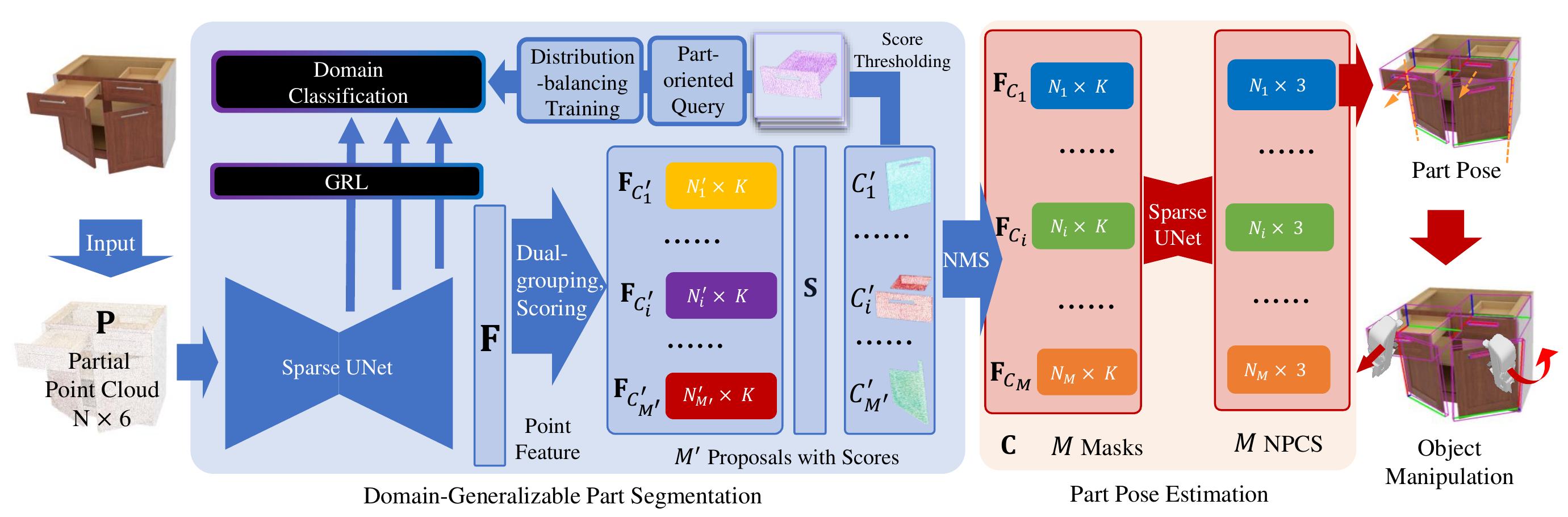}
\caption{\textbf{An Overview of Our Domain-generalizable Part Segmentation and Pose Estimation Method.} We introduce a part-oriented domain adversarial training strategy that can tackle multi-resolution features and distribution imbalance for the domain-invariant GAPart feature extraction. The training strategy tackles the challenges in our tasks and dataset, significantly improving the generalizability of our method for part segmentation and pose estimation. }
\label{fig:pipe}
\end{figure*}

\section{Problem Formulation}
\label{sec:prob_def}
Given the GAPart definition and the proposed GAPartNet dataset, we investigate the problems of cross-category generalizable object perception and manipulation.

\paragraph{Perception.} The input to our pipeline is a \textit{partial} colored point cloud observation of the object $ \mathbf{P} \in  \mathbb{R}^{N\times 3}$, where $N$ denotes the number of points.
Assume the object contains $L$ GAParts and the $i$-th part is with a class label $p_i \in \{1, ..., 9\}$.
Then the goal for perception is as follows:
for each individual GAPart, locating its segmentation masks $C_{i}$ and recognizing its part pose, \ie, a rotation $\mathbf{R}_i \in \text{SO}(3)$, a translation $\mathbf{t}_i \in \mathbb{R}^{3}$, and a size $\mathbf{s}_i \in \mathbb{R}^{3}$.

Note that the perception tasks are carried out in a cross-category, domain-generalizable fashion, \ie, the perception network is trained on objects from a set of seen object categories $\{O^S_j\}_j$ (\ie, seen domains $\{D^S_j\}_j$), and is expected to generalize to unseen object categories $\{O^U_j\}_j$ (\ie, unseen domains $ \{D^U_j\}_j$).

\paragraph{Manipulation.} We need to develop a pose-based interaction policy $\pi$ for generalizable part-based object manipulation. Given a single partial point cloud observation $\mathbf{P} \in \mathbb{R}^{N \times 3}$, the robot needs to manipulate the target part using the previous understanding for the GAParts, \eg, open a door on an object from a previous unseen object category.


\section{Method}
\label{sec:method}
Our proposed pipeline for domain-generalizable 3D part segmentation and pose estimation is shown in \cref{fig:pipe}.

\subsection{Domain-generalizable 3D Part Segmentation}

\paragraph{Architecture Overview.} 
Following the previous works \cite{jiang2020pointgroup, vu2022softgroup}, with the input point cloud $\mathbf{P}$, our 3D part segmentation network leverages a Sparse UNet \cite{graham2017submanifold} as the backbone to extract point-wise feature $ \mathbf{F} $ with $K$ channels, followed by a \textit{Dual Set Grouping} module introduced by \cite{jiang2020pointgroup} to generate $M^\prime$ mask proposals $ \mathcal{C}^\prime = \{C^\prime_1, C^\prime_2,\dots,C^\prime_{M^\prime}\}$. The proposals are then passed through a \textit{Scoring} module which predicts confidence scores $\mathbf{S}$, with $\mathbf{S}_i$ as the score for the proposal $i$, followed by Non-Maximum Suppression (NMS) to output final $M$ segmentation masks $ \mathcal{C}=\{C_1, C_2, \dots ,C_M\} $. 
Most importantly, to enable domain-invariant feature extraction for mask proposals and tackle the aforementioned challenges, we introduce a domain adversarial training strategy for 3D part segmentation to help learn domain-invariant features.

\paragraph{Domain-invariant GAPart Feature Learning.} Inspired by \cite{ganin2016domain,ganin2015unsupervised,li2018domain}
, we introduce a domain classifier $ \mathcal{D} $ and a Gradient Reverse Layer (GRL) at the training time for domain adversarial training, as shown in \cref{fig:pipe}. Specifically, the classifier $ \mathcal{D} $ takes the features as input and tries to distinguish the different domains, while the GRL passes the negative gradients of the classification back to the feature extractor, which encourages domain-invariant feature extraction during this adversarial training procedure. 

Furthermore, to address the challenges mentioned in \cref{sec:prob_def}, we consider 1) how to process the feature from the part segmentation pipeline to make the GAPart feature domain-invariant, 2) where to place the domain classifier to better tackle parts with different sizes, and 3) how to do domain adversarial training to deal with the distribution-imbalance. The designed techniques are as follows.

1) Part-oriented Feature Query ($\mathbf{Q}$). 
To better handle the huge variations in part contexts across different domains, the part features need to be context-invariant and contain less domain-relative information. An intuitive design is to make the domain classifier $ \mathcal{D} $ part-oriented (\ie, taking foreground part features as input and domain labels as output), which can help the feature extractor focus on the foreground (\ie, the GAParts) rather than the background (\ie, the rest of the object bodies). Specifically, we query the features of mask proposals $\{\mathbf{F}_{C_i^\prime}\}_{i}$ with scores above the threshold $\mathbf{s}_{thre}$ from the feature $\mathbf{F}$ and pass them to the domain classifier. The domain discrimination loss is
$$
\mathcal{L}_{\mathbf{Q}-\text{adv}}(\mathbf{F}) = \frac{1}{M_s^\prime}\sum\limits_{i=1}^{M^\prime} \mathbbm{1}_{\{\mathbf{S}_i > \mathbf{s}_{thre}\}}\mathcal{L}_\text{cls}^\text{adv}(
\mathcal{D}(\mathbf{F}_{C_i^\prime}), d_i), 
$$
\noindent where $M_s^\prime$ indicates the number of proposals with scores above the threshold, $d_i$ is the domain label (\ie, object category) of the mask proposal $C_i^\prime$, and $\mathcal{L}_\text{cls}^\text{adv}(\cdot,\cdot)$ denotes the domain classification loss.

2) Multi-resolution ($\mathbf{R}$). Part instances come in significantly different sizes, \eg, a door can be an order of magnitude larger than a handle. We thus propose to extract the mask proposal features from different UNet layers in different resolutions, so that the size variances of GAParts can be taken care of. In the implementation, we choose three hidden layers from the UNet decoder and query proposal features from the three features $\{\mathbf{F}^{l}\}_{l}$ respectively.

Combined with multi-resolution,  $\mathcal{L}_{\mathbf{Q}-\text{adv}}$ can be re-written as follows:
$$\mathcal{L}_{\mathbf{QR}-\text{adv}}(\{\mathbf{F}^{l}\}_{l}) = \sum\limits_{l = 1}^3 w_l \mathcal{L}_{\mathbf{Q}-\text{adv}}(\mathbf{F}^l),
$$
 \noindent where $\mathcal{L}_{\mathbf{Q}-\text{adv}}(\mathbf{F}^l)$ indicates the domain discrimination loss for features queried from the $l^\text{th}$ layer and $w_l$ is the corresponding weight for each layer.

Note that these multi-resolution features only serve domain-adversarial learning for parts with different sizes and are not involved in the grouping for mask proposals.

3) Distribution-balancing ($\mathbf{B}$).
\label{focal}
As is often the case in the real world, part instances on different objects can be extremely imbalanced. We thus introduce a part-level domain discrimination focal loss inspired by \cite{Lin_2017_ICCV} for adversarial training to tackle this problem. 
Combining with distribution-balancing, $\mathcal{L}_{\mathbf{Q}-\text{adv}}$ can be modified as follows:
$$
\mathcal{L}_{\mathbf{QB}-\text{adv}}(\mathbf{F}) =  \frac{1}{M_s^\prime}\sum\limits_{i=1}^{M^\prime}  \mathbbm{1}_{\{\mathbf{S}_i > \mathbf{s}_{thre}\}}w_{d_i}^{p_i}\mathcal{L}_\text{cls}^\text{adv}(
\mathcal{D}(\mathbf{F}_{C_i^\prime}), d_i),
$$
$$w_{d_i}^{p_i} =  -\alpha_{d_i}^{p_i} (1-acc_{d_i}^{p_i})^{\gamma},
$$ 
\noindent where for the specific part class $p_i$ and the domain $d_i$ of a proposal $C^\prime_i$, the loss weight $w_{p_i}^{d_i}$ is determined by a hyper-parameter $\alpha_{d_i}^{p_i}$, negatively correlated with the domain distribution, $acc_{d_i}^{p_i}$, the mean accuracy of the classification for the domain $d_i$ in the part class $p_i$, and $\gamma$, a hyper-parameter.

With the three techniques introduced, our proposed domain adversarial training method is part-oriented and can tackle multi-resolution features as well as distribution imbalance, which better encourages domain-invariant GAPart feature learning. 
The final domain adversarial loss is
$$\mathcal{L}_{\mathbf{QRB}-\text{adv}}(\{\mathbf{F}^{l}\}_{l}) = \sum\limits_{l = 1}^3 w_l \mathcal{L}_{\mathbf{QB}-\text{adv}}(\mathbf{F}^l),
$$ 
and the total loss for domain-generalizable part segmentation is as follows:
$$\mathcal{L}_\text{seg}^\text{DG} = \mathcal{L}_\text{seg} + \mathcal{L}_{\mathbf{QRB}-\text{adv}},
$$
where $\mathcal{L}_\text{seg}$ is the part segmentation loss without domain adversarial training.

\subsection{Part Pose Estimation}
\label{sec:dg_pose}

\paragraph{NPCS Map Prediction and Pose Fitting.} 
\label{sec: NPCS}
For each predicted part segmentation mask $C_i$, we query its mask feature ${\mathbf{F}}_{C_i}$ from the feature $\mathbf{F}$. Then the $NPCS$-$Net$ is used for point-wise NPCS coordinates regression. Applying RANSAC \cite{fischler1981random} for outlier removal and Umeyama algorithm \cite{umeyama1991least}, we estimate the 7-dimensional rigid transformation and obtain the pose of the predicted part.
Based on the domain-invariant feature, thanks to our domain adversarial training, the prediction of NPCS values in our pipeline can be independent of the context, color, etc. of the part. This can significantly improve the generalizability of our part pose estimation method. 

\paragraph{Symmetry-aware Pose Estimation and Joint Prediction.}
\label{sec:Sym&Joint}
To tackle the symmetries naturally existing in some GAPart classes, we design a symmetry-aware NPCS regression loss that can tolerate different symmetry patterns for different part classes. We then follow our GAPart pose definition to simplify the joint prediction procedure. For each GAPart class, the part pose definition contains a wealth of information, including the joint position and direction, where we can directly get the joint position and direction instead of relying on an additional network for estimation like~\cite{li2020category}.

\subsection{Interaction Policy}
\label{sec:interaction}

Given part segmentation and pose estimation, based on the proposed GAPart pose definition where actionability information is included, we design part-pose-based, effective interaction policies for part-based object manipulation, which provide the community with a novel approach to cross-category robotic manipulation and interaction tasks. 

More design and implementation details of our method can be found in the appendix.

\begin{table*}[ht]
\centering
\renewcommand\arraystretch{0.9}
\resizebox{.99\textwidth}{!}{
\begin{tabular}
{c|c|ccccccccccc}
\hline
&& Ln.F.Hl. & Rd.F.Hl. & Hg.Hl. & Hg.Ld.& Sd.Ld.  & Sd.Bn  & Sd.Dw. & Hg.Dr. & Hg.Kb. &Avg.AP
& Avg.AP50
\\ \hline
\multirow{4}{*}{Seen (\%)}   & PG\cite{jiang2020pointgroup} &86.1 &23.0 &84.6 &80.01&88.3 &49.3 &62.6 &92.8 &34.6 &57.3 &66.8 \\ 
                        & SG\cite{vu2022softgroup} & 57.8 & \textbf{93.6} & 81.2 & 76.0 & 89.3 & 25.2 & 50.8 & 93.9 & 51.5 & 58.5 & 68.8 \\ 
                        & AGP\cite{liu2022autogpart} &86.8 &20.3  &87.7 &79.7  &89.4  &62.3   &61.6  &92.5  & 16.7 & 57.2 &66.3 \\ 
                        & Ours       &\textbf{89.2} &54.9 &\textbf{90.4} &\textbf{84.8} &\textbf{89.8} &\textbf{66.7} &\textbf{67.2} &\textbf{94.7} &\textbf{52.9} &\textbf{67.6} &\textbf{76.5} \\ \hline
\multirow{4}{*}{Unseen (\%)} & PG\cite{jiang2020pointgroup} &32.44 &\textbf{}9.8 &2.1 &26.8 &0.0 &42.6 &57.0 &63.9 &1.7 &21.9 &26.3 \\ 
                        & SG\cite{vu2022softgroup} & 25.8 & 5.0 & 0.4 & 33.9 & 0.6 & \textbf{51.5} & 51.2 & \textbf{}69.0 & 12.1 & 22.0 & 27.7 \\ 
                        & AGP\cite{liu2022autogpart} &\textbf{45.6} &4.8  &\textbf{3.1}  &34.3 &0.0  &47.8 &64.1 &63.1  &11.5  &25.7  &30.5  \\ 
                        & Ours       & {\textbf{45.6}} & {\textbf{40.0}} & {\textbf{3.1}} & {\textbf{40.2}} & {\textbf{5.0}} & \multicolumn{1}{c}{\textbf{}49.1} & \multicolumn{1}{c}{\textbf{64.2}} &\textbf{69.1} &\textbf{23.4}&\textbf{32.0}  &\textbf{37.2}  \\ \hline
\end{tabular}
}
\caption{\textbf{Results of Part Segmentation on Seen Object Categories and Unseen Object Categories in terms of Per-part-class AP50 (\%), Average AP50 (\%), and Average AP (\%).} Ln.=Line. F.=Fixed. Rd.=Round. Hl.=Handle. Ld.=Lid. Bn.=Button. Dw.=Drawer. Dr.=Door. Kb.=Knob.  PG=PointGroup\cite{jiang2020pointgroup}. SG=SoftGroup\cite{vu2022softgroup}.
AGP=baseline modified from AutoGPart\cite{liu2022autogpart}.}
\label{table:mainres}
\end{table*}

\begin{table}[ht]
\scriptsize
\renewcommand\arraystretch{0.9}
\centering
\setlength{\tabcolsep}{0.5em}
\resizebox{.47\textwidth}{!}{
\begin{tabular}{cccc|cccc}
\hline
&   &   &   & \multicolumn{2}{c|}{Seen (\%)}   & \multicolumn{2}{c}{Unseen (\%)}                              \\ \hline
\begin{tabular}[c]{@{}c@{}}use\\ $\text{adv}$\end{tabular} & \begin{tabular}[c]{@{}c@{}}use\\ $\mathbf{Q}$-adv\end{tabular} & \begin{tabular}[c]{@{}c@{}}use\\ $\mathbf{R}$-adv\end{tabular} & \begin{tabular}[c]{@{}c@{}}use\\ $\mathbf{B}$-adv\end{tabular} & \begin{tabular}[c]{@{}c@{}}Avg. \\ AP\end{tabular} & \multicolumn{1}{c|}{\begin{tabular}[c]{@{}c@{}}Avg. \\ AP50\end{tabular}} & \begin{tabular}[c]{@{}c@{}}Avg. \\ AP\end{tabular} & \begin{tabular}[c]{@{}c@{}}Avg. \\ AP50\end{tabular} \\ \hline
\XSolidBrush&         &           &  &61.1 & \multicolumn{1}{c|}{71.1} &22.2  &28.1    \\
\Checkmark&           &           &  &61.0 & \multicolumn{1}{c|}{70.6} &23.2  &29.8    \\
\Checkmark& \Checkmark&           &  &62.8 & \multicolumn{1}{c|}{71.6} &27.1  &32.3    \\
\Checkmark& \Checkmark& \Checkmark&  &64.9 & \multicolumn{1}{c|}{73.7} &29.6  &35.0    \\
\Checkmark& \Checkmark& \Checkmark& \Checkmark&\textbf{67.6}   & \multicolumn{1}{c|}{\textbf{76.5}}   &\textbf{32.0}  &\textbf{37.2}    \\ \hline
\end{tabular}
}
\caption{\textbf{Ablation Studies for Domain-generalizable Part Segmentation.} The left four columns stand for 
using adversarial learning, part-oriented feature query technique, multi-resolution technique, distribution-balancing technique, respectively.
}
\label{table:ablation}
\end{table}

\begin{table}[t]
\centering
\setlength{\tabcolsep}{0.2em}
\resizebox{.47\textwidth}{!}
{
\begin{tabular}{c|c|cccccccc}
\hline
    &  & {\boldmath{$R_{e}$}} $\downarrow$ & {\boldmath{$T_{e}$}}$\downarrow$ & {\boldmath{$S_{e}$}}$\downarrow$ & {\boldmath{$\theta_{e}$}}$\downarrow$ & {\boldmath{$d_{e}$}}$\downarrow$ & \textbf{mIoU} $\uparrow$ & $\textbf{A}_{5}$ $\uparrow$ & $\textbf{A}_{10}$ $\uparrow$\\ \hline
\multirow{3}*{Seen} 
&PG\cite{jiang2020pointgroup}  
& 14.3&0.034&0.039&7.947&0.020&49.4&24.4&47.0\\ 
&AGP\cite{liu2022autogpart}        
&14.4&0.036&0.039&7.955&0.021&48.7&26.8&49.1  \\
&Ours             
&\textbf{9.9}&\textbf{0.024}&\textbf{0.035}&\textbf{7.4}&\textbf{0.014}&\textbf{51.2}&\textbf{28.3}&\textbf{53.1}\\ \hline
\multirow{3}*{Unseen} 
&PG\cite{jiang2020pointgroup}        
&18.2&\textbf{}0.056&0.073&12.0&0.031&36.2&19.2&42.9  \\
&AGP\cite{liu2022autogpart}        
&18.2&0.57&0.076&11.9&0.029&36.3&20.8&46.5  \\
&Ours             
&\textbf{14.8}&\textbf{0.047}&\textbf{0.067}&\textbf{11.3}&\textbf{0.024}&\textbf{40.6}&\textbf{23.4}&\textbf{51.6}\\ \hline
\end{tabular}
}
\caption{\textbf{Results of Part Pose Estimation on Seen and Unseen Object Categories in terms of \boldmath{$R_{e}$} ($^{\circ}$), \boldmath{$T_{e}$} (cm), \boldmath{$S_{e}$} (cm), \boldmath{$\theta_{e}$} ($^{\circ}$), \boldmath{$d_{e}$} (cm), \textbf{mIoU}=\textbf{3D mIoU} (\%), $\textbf{A}_5$=\textbf{5$^{\circ}$5cm accuracy} ($\%$), $\textbf{A}_{10}$=\textbf{10$^{\circ}$10cm accuracy} ($\%$).} PG=baseline modified from PointGroup \cite{jiang2020pointgroup}. AGP=baseline modified from AutoGPart\cite{liu2022autogpart}.} 
\label{table:PoseEstimation}
\end{table}

\section{Experiments}
\subsection{Data Preparation}
With our dataset described in \cref{sec:dataset_def}, we render RGB-D images of objects with annotations using the SAPIEN environment \cite{xiang2020sapien} and obtain point cloud observations from back-projection. 
To study the cross-category generalizability of our method, we split the 27 object categories into 17 seen and 10 unseen categories, ensuring that all GAPart classes exist in both seen and unseen object categories. We train the network on seen categories and evaluate its GAPart understanding on unseen categories. 

\subsection{Cross-category Part Segmentation}

\paragraph{Evaluation Metrics.}
Following the previous 3D semantic instance segmentation benchmarks in ScanNetV2 \cite{dai2017scannet} and S3DIS \cite{armeni_cvpr16},  we use the widely-adopted metric average precision to evaluate the performance of part segmentation. Specifically, AP50, the average precision with Intersection over Union (IoU) threshold 50\%, is used to evaluate the performance on each part class and the overall performance. As a complementary, we also use AP, the average precision averages over IoU thresholds from 50\% to 95\% with a step of 5\%, to evaluate the overall performance.


\paragraph{Main Results.}
\cref{table:mainres} shows the quantitative comparisons between our method and previous state-of-the-art methods of 3D semantic instance segmentation (\ie, PointGroup\cite{jiang2020pointgroup}, SoftGroup\cite{vu2022softgroup}). We also set up a baseline modified from AutoGPart\cite{liu2022autogpart}, whose task is different from ours thus we directly combine their methods with the original PointGroup\cite{jiang2020pointgroup} pipeline for comparison. 

In both seen and unseen object categories, our method shows significant improvement compared to the others. For AP50, our method achieves 76.5\% in seen categories, which beats the second-runner by absolutely 7.7\% and relatively 11.2\%. In unseen categories, our method achieves 37.2\%, absolutely 6.7\% and relatively 22.0\% better than the second-runner, which shows significant relative improvement in unseen categories. It shows that our method could extract better domain-invariant features for parts and thus have great generalizability across object categories.

\begin{figure}[t]
\centering
\includegraphics[width=1\linewidth]{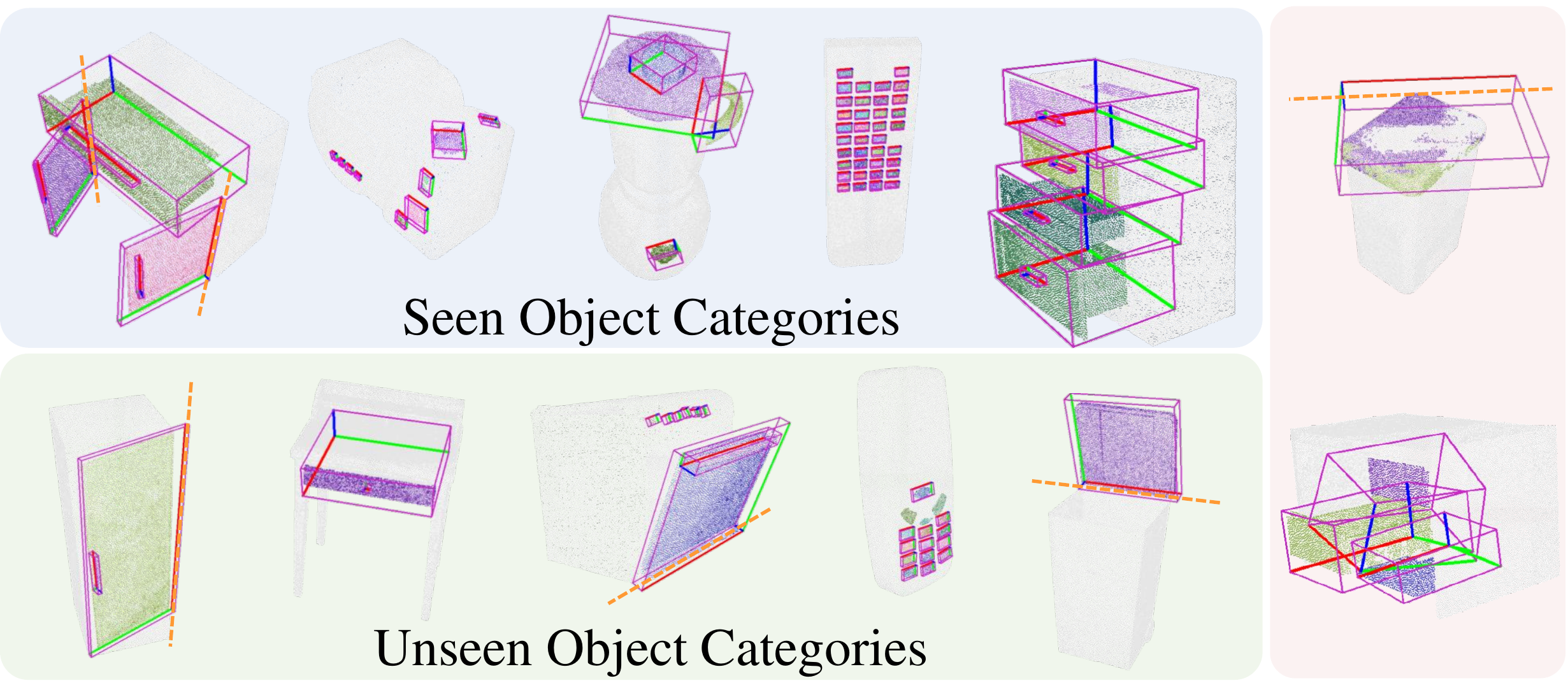}
\caption{\textbf{Qualitative Results of Perception.} Left two figures show the results of cross-category part segmentation and pose estimation on seen and unseen categories, while the right shows failure cases. Here we only show the revolute joint estimation results.}
\label{fig:gallery_vis}
\end{figure}

\paragraph{Ablation Studies.} 
We conduct sufficient comparisons to demonstrate that our techniques contribute significantly to the generalizability across object categories, as shown in Table \ref{table:ablation}.
Comparing the top two rows, we show that the domain adversarial training with the object global features as input helps the generalization to unseen categories, but somewhat at the expense of performance in seen categories. With our part-oriented feature query technique (rows 2,3), the performance improves no matter on seen or unseen categories. The multi-resolution technique also contributes to the performance in the two areas (rows 3,4). 
The distribution-balancing technique (rows 4,5) takes the 
performance of our method a step further and achieves strong precision and generalizability.

\subsection{Cross-Category Part Pose Estimation}
\label{sec:pose_exp}

\paragraph{Evaluation Metrics.}
We use the following metrics to evaluate the performance of part pose estimation: 
{\boldmath{$R_{e}$}} ($^{\circ}$), average rotation error;
{\boldmath{$T_{e}$}} (cm), average translation error;
{\boldmath{$S_{e}$}} (cm), average scale error;
{\boldmath{$\theta_{e}$}} ($^{\circ}$) average rotation error of part interaction axis;
{\boldmath{$d_{e}$}}(cm) average translation error of part interaction axis;
\textbf{3D mIoU} ($\%$), the average 3D IoU between ground-truths and predicted bounding boxes;
 \textbf{5$^{\circ}$5cm accuracy} ($\%$), the percentage of pose predictions with rotation error $<$ 5$^{\circ}$ and translation error $<$ 5cm;
 \textbf{10$^{\circ}$10cm accuracy} ($\%$), the percentage of pose predictions with rotation error $<$ 10$^{\circ}$ and translation error $<$ 10cm.
We evaluate part pose only when the part is detected.

\paragraph{Main Results.} \cref{table:PoseEstimation} shows the results of our method and the baselines on part pose estimation. We modify PointGroup \cite{jiang2020pointgroup} and AutoGPart \cite{liu2022autogpart} as baselines. Our method outperforms the baselines on most of the metrics in seen categories, and on all of the metrics in unseen categories, which shows the value of our domain-invariant feature extraction. With our domain adversarial training strategy and the three techniques introduced, the performance of part pose estimation improves a lot, especially in unseen categories. Qualitative results are shown in \cref{fig:gallery_vis}.

\subsection{Cross-category Part-based Manipulation}
We showcase the usefulness of the concept of GAPart by performing cross-category, part-based object manipulation on four basic tasks. We use SAPIEN \cite{xiang2020sapien} environment for simulation and set up four tasks based on SAPIEN Manipulation Skill Benchmark \cite{mu2021maniskill}, \ie, opening drawers, opening doors, using handles, pressing buttons.

\paragraph{Task Setting.}  These four tasks exemplify robot manipulation under the motion constraint of a prismatic or a revolute joint, where a gripper is used on seen and unseen categories. The success of object manipulation is defined as opening up the target part for 90\% of the motion range within 1,000 time-steps. and coming to a stable stop at the end.

\begin{figure}[t]
\centering
\includegraphics[width=1\linewidth]{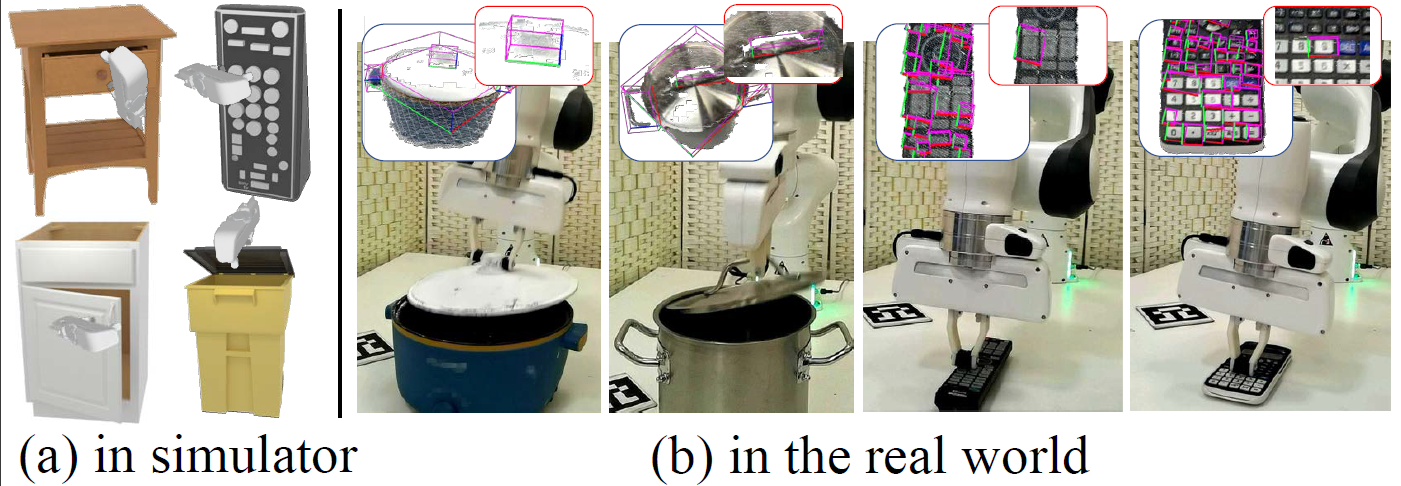}
\caption{\textbf{Part-based Object Manipulation.} Left (a): In the simulator. Right (b): In the real world. For each subfigure in (b), the perception result is shown in the left box, while the target part is shown in the right box.}
\label{fig:result}
\end{figure}

\paragraph{Heuristics Design and Experiments in the Simulator.} We first do cross-category part segmentation and pose estimation using our perception method. Based on the predictions of the part poses, we move the robot arm toward the target part, turn the gripper in the direction suitable for grabbing, and then close the gripper. Finally, we move the gripper along the proposed trajectories toward the target position, following our GAPart pose definition. The results show that our perception model and manipulation heuristics can work well, achieve good performance on these tasks, and generalize to objects from unseen categories. Exemplar results are shown in~\cref{fig:result} (a). 

\paragraph{Real-world Experiments.} Although trained on synthetic data, our method can be used in the real world. Experiments show that our method can successfully predict part segmentation and poses on real objects. We further show that cross-category part-based object manipulation can be successfully performed by robot arms using our method, as shown in~\cref{fig:result} (b). 

 More experiment details, quantitative and qualitative results can be found in the appendix.

\section{Conclusion}

In this work, we reason that learning generalizable and actionable parts is the key to an intelligent agent capable of cross-category object perception and manipulation. We introduce the concept of GAPart and present the GAPartNet dataset by annotating cross-category part semantics and poses. We explore three cross-category tasks based on GAParts: part segmentation, part pose estimation, and part-based object manipulation. Our proposed approach, adopting a domain generalization perspective, outperforms previous works in segmentation and pose estimation. Furthermore, we design part-pose-based interaction policies that enable effective and generalizable object manipulation in both the simulator and the real world, thanks to our GAPart definition and our domain-generalizable perception model.

\paragraph{Limitations.} The cross-category tasks are challenging, and there is still room for improvement in generalizability. Our heuristic method for object manipulation relies on precise part pose predictions, which is an area for future research to achieve more robust manipulation strategies.
More discussions can be found in the appendix.

\section{Acknowledgements}
This work is supported in part by the National Key R\&D Program of China (2022ZD0114900).

{\small
\bibliographystyle{ieee_fullname}
\bibliography{egbib}
}

\clearpage
\appendix
\section{Dataset and Data Annotation}

\subsection{Data Annotation}

To construct a large-scale part-centric interactive dataset, great effort is needed to clean up and annotate existing object shapes. We first identify the issues with the existing database, and then we develop a systemic pipeline for annotating the large-scale dataset.

\textbf{Data sources.} GAPartNet dataset is constructed based on two existing datasets, PartNet-Mobility~\cite{xiang2020sapien} and AKB-48~\cite{liu2022akb}. Focusing on the GAParts we define, we select 23 object categories from PartNet-Mobility and 4 object categories from AKB-48. Most of the 3D object shapes in GAPartNet are from PartNet-Mobility. Since the texture of shapes in PartNet-Mobility is all synthetic, to mitigate the sim-to-real gap, we further leverage the shapes from AKB-48 whose texture is scanned from the real world.

Note that both PartNet-Mobility and AKB-48 have the object categories \textit{Box, Bucket, TrashCan}. Although they use the same category names, their shapes can be very different. A TrashCan from PartNet-Mobility and one from AKB-48 can have significant differences in geometry, the same as Box and Bucket. So we do not merge them together into one object category but keep their original categories.

\textbf{Issues with Existing Database. }The original PartNet-Mobility~\cite{xiang2020sapien} and AKB-48~\cite{liu2022akb} lack of directly usable information we need for our new annotations. First, they do not provide directly usable consistent semantic annotations to similar parts across object categories. For example, some handles on \textit{Door} are labeled as \textit{door}, while some doors on \textit{StorageFurniture} are labeled as \textit{frame}. Secondly, their original annotations are not as fine-grained as we need. Specifically, fixed handles, \ie line fixed handles and round fixed handles, are not annotated as individual parts, since they are attached to either base bodies or other movable parts. Their meshes are merged with others which leaves rare semantic cues to re-separate them. Finally, there are a lot of meshes of parts that we care about are imperfect, which seriously limits either the quality of our pose annotations or the quality of rendered images.

\textbf{Data Annotation Effort.} To address these issues, we first manually go over all objects to re-separate the meshes of fixed handles from the original 3D object shapes. We also modify the kinematic chains to re-merge these meshes into new links and add corresponding fixed joints, which provides more consistent annotations and is beneficial for following robotic tasks. In this step, more than 1,000 fixed handles are re-separated and re-merged. Secondly, we go over all 1,166 objects in GAPartNet and clean all original semantic annotations to align with our GAPart class definition. Thirdly, we manually use MeshLab and some heuristics to modify imperfect meshes, not only cutting the redundant meshes off but also fixing the one-sided meshes to facilitate the annotating and rendering. In this step, more than 100 object instances are modified.

Finally, with the 1,166 3D object shapes with new semantic annotations and modified meshes, we use a lot of heuristics to fit the oriented tight bounding boxes of all 8,489 GAParts, corresponding to their canonical orientations, and add our pose annotations. With our effort, GAPartNet is capable of detection, segmentation, pose estimation, and manipulation on cross-category generalizable and actionable parts.

\subsection{Dataset Rendering}

We use the SAPIEN 2.0 environment~\cite{xiang2020sapien} to render a large-scale dataset from our GAPartNet objects, consisting of partial point clouds, part semantic segmentation masks, part instance segmentation masks, NPCS maps, and part pose annotations, which covers all the data needed for the proposed part segmentation, part pose estimation and part-based object manipulation tasks.

\textbf{Environment Settings.} We turn on the ray-tracing mode of SAPIEN to get more sense of reality. During rendering, we randomize the joints' poses of the articulated objects and randomly pick a camera position within a reasonable perspective. Specifically, we manually set the range of camera position for each object category to get desirable views of each object, making sure we do not look at the back of a StorageFurniture, or from beneath an Oven, neither from too far nor too close. In the meantime, we randomly dim the ambient light within [10\%, 90\%] and randomly rotate the camera within $\pm 5^\circ$.

The output image resolution is set to $800 \times 800$. For each object, we render 32 RGB images. Along with each RGB image, we also obtain the segmentation masks and the depth image using built-in features of the SAPIEN environment. Additionally, we compute NPCS maps and oriented tight bounding boxes as part pose annotations for all GAParts.

\textbf{Point Cloud Sampling.} Using camera intrinsics, 2D RGB images, and depth images, we do back-projection to obtain dense, partial point clouds. We sample 20,000 points for each dense point cloud using Farthest-Point-Sampling (FPS). While sampling the point clouds, we also generate corresponding ground truth of semantic segmentation, instance segmentation, and NPCS maps. These 20,000-point point clouds and their annotations are computed offline for speeding up the following 3D tasks.

\section{More Details on Part Segmentation and Part Pose Estimation}

\subsection{Details on Network Architecture}

\textbf{Architecture.}
The vision network has a similar architecture as PointGroup \cite{jiang2020pointgroup}. Please refer to the original PointGroup paper for details. In our work, we set the cluster radius to 0.03 and the cluster point number threshold to 5 to get good segmentation results in the GAPartNet dataset. 
The input point cloud $\mathbf{P}$ is first voxelized into a $100\times100\times100$ voxel grid. The backbone UNet consists of an encoder and a decoder, both with a depth of 7 (with channels of [16, 32, 48, 64, 80, 96, 112]), and outputs a point-wise feature $\mathbf{F}$ with $K$ channels, where $K = 16$. After grouping, each mask proposal $C'_i$ is normalized and voxelized again into a $50\times50\times50$ voxel grid and passed through the $Scoring$ module, which consists of a 2-depth UNet (with channels of [16, 32]) for point-wise feature extraction, an ROI Pooling layer for foreground feature merging, and a linear layer for confidence score $\mathbf{S}_i$ prediction. During inference, points with binary classification scores below 0.4 are filtered out as background, and proposals with fewer than 5 points or a score lower than 0.09 are discarded. Finally, Non-Maximum Suppression (NMS) with an IoU (Intersection over Union) threshold of 0.3 is applied to get the final segmentation masks $\mathcal{C}$.

For domain adversarial learning, we introduce a Gradient Reverse Layer (GRL) with $\alpha = 0.3$ for the negative gradients and three domain discriminators with similar architectures as the $Scoring$ module mentioned above for domain classification. We place the three discriminators at the $2$-nd, $4$-th, and $6$-th decoder layers of the backbone UNet, so the three discriminators can take different features from the three layers of the backbone for domain classification. Each discriminator takes the queried points and the corresponding features as input and predicts the domain labels. The domain discriminators are only used during the training procedure, and the proportion of classification is set to 0.05 in our implementation.

For each segmentation mask $C_i$, the point-wise feature $\mathbf{F}_{C_i}$ queried from $\mathbf{F}$ is passed through the $NPCS$-$Net$, consisting of a 2-depth UNet (with channels of [16, 32]) and three Multilayer Perceptrons (3-MLP) for poise-wise NPCS prediction. Note that in practice, we use 9 different groups of 3-MLP to predict NPCS coordinates in 9 channels, and we only supervise the channel corresponding to the ground truth semantic label.

\subsection{Details on Supervision}

\textbf{Symmetry-aware Part Pose Estimation.}
Since each part class in GAPart has different symmetry patterns, they should be handled case by case. We design the symmetry-aware NPCS loss as follows:

\textit{Type 1} (\ie, line fixed handle, hinge handle):
we tolerate the $180^\circ$ symmetry along the $z$ axis for this symmetry type.

\textit{Type 2} (\ie, hinge door, hinge lid):
we tolerate the $180^\circ$ symmetry along the $y$ axis for this symmetry type.

\textit{Type 3} (\ie, slider button, slider lid, round fixed handle):
we tolerate the rotation along the $z$ axis and flipping along the $x$-$y$ plane for this symmetry type. In our implementation, we split the continuous rotation angles into 12 discrete angles for supervision.

\textit{Type 4} (\ie, hinge knob):
we tolerate the rotation along the $z$ axis for this symmetry type. In our implementation, we split the continuous rotation angles into 12 discrete angles for supervision.

\textit{Type 5} (No symmetry, \ie, slider drawer) :
we do not tolerate any symmetry for this symmetry type.

The design of NPCS loss $\mathcal{L}_\text{NPCS}$ is similar to~\cite{wang2019normalized}. We use soft-L1 loss and for each tolerated symmetry pattern, we supervise the minimal loss in the set. For more implementation details, please refer to~\cite{wang2019normalized}.

\textbf{Loss Function.} 
The whole training procedure of the network can be divided into four stages.

For the first stage (0-5 epochs), we only supervise the semantic prediction and the offset prediction branches with the same loss functions $\mathcal{L}_\text{sem}$ and $\mathcal{L}_\text{off}$ as PointGroup~\cite{jiang2020pointgroup}. Please refer to \cite{jiang2020pointgroup} for more details. 

For the second stage (5-10 epochs), we add the score loss $\mathcal{L}_\text{sco}$ for the proposals' IoU prediction, following the design of \cite{jiang2020pointgroup}.

For the third stage (10-15 epochs), we add the symmetry-aware NPCS loss $\mathcal{L}_\text{NPCS}$ for the NPCS prediction, as introduced above.

For the fourth stage (after 15 epochs), we introduce our domain adversarial learning strategy after the part segmentation network can output good proposals and corresponding proposal scores, similar to~\cite{ganin2016domain}. The total loss in this stage can be formulated as
$$
\mathcal{L}
= \mathcal{L}_\text{\textbf{QRB}-adv} + \mathcal{L}_\text{sem} + \mathcal{L}_\text{off}+\mathcal{L}_\text{sco}+\mathcal{L}_\text{NPCS},
$$
where $\mathcal{L}_\text{\textbf{QRB}-adv}$ denotes the domain adversarial loss.

\subsection{Details on Pose Fitting and Joint Prediction}

\textbf{Pose Fitting.} 
Given a predicted 3D part mask with its NPCS map, we use RANSAC~\cite{fischler1981random} for outlier removal and Umeyama
algorithm~\cite{umeyama1991least} to estimate the 7-dimensional rigid transformation.

\textbf{Joint Parameter Prediction.} 
We simplify the joint parameter prediction process thanks to the unified definition of our GAParts. After estimating the bounding box for each part, we can leverage the definition of the GAPart to directly calculate the joint parameters. For example, given the bounding box of a slider button, we can directly query its prismatic joint parameter, which is along the $z$ axis in the part canonical space.

\subsection{Training Procedure}
Our model is trained in an end-to-end manner with maximum training epochs of 200. We use the Adam optimizer with a batch size of 32 and a learning rate of 0.001. The whole training procedure takes around 1.5 days on a single NVIDIA GeForce RTX 2080 Ti GPU. Note that the domain adversarial training is very unstable, we thus use five seeds to train it and select the best one. What's more, to boost performance, we progressively use the multi-resolution training strategy, which improves performance.

\subsection{Seen/Unseen Object Categories Splitting}

\textbf{17 Seen Categories.} Box, Bucket, Camera, CoffeeMachine, Dishwasher, Keyboard, Microwave, Printer, Remote, StorageFurniture, Toaster, Toilet, WashingMachine, Bucket (AKB-48), Box (AKB-48), Drawer (AKB-48), Trashcan (AKB-48).

\textbf{10 Unseen Categories.} Door, KitchenPot, Laptop, Oven, Phone, Refrigerator, Safe, Suitcase, Table, TrashCan.

\subsection{Baseline Experiments}
\textbf{PointGroup\cite{jiang2020pointgroup}.} The PointGroup baseline is modified from \cite{jiang2020pointgroup}. We add our NPCS prediction branch to the vanilla PointGroup. The final loss can be formulated as $\mathcal{L}_\text{PointGroup} = \mathcal{L}_\text{seg} + \mathcal{L}_\text{NPCS}$, where $\mathcal{L}_\text{NPCS}$ is the same as our method.

\textbf{AutoGPart\cite{liu2022autogpart}.} Following AutoGPart\cite{liu2022autogpart},  we introduce a similar intermediate supervision for generalizable part segmentation. We build a parametric supervision model $\mathcal{M(\cdot|\theta)}$ to find a proper intermediate part segmentation supervision, which can be learned through a  ``propose, evaluate, update" strategy. We use each object category as each ``sub-domain" in AutoGPart and use the same hyper-parameters for the intermediate auxiliary loss. We still add our NPCS prediction branch to the network for part pose estimation. The final loss can be formulated as $\mathcal{L}_\text{AGP} = \mathcal{L}_\text{seg} + \mathcal{L}_\text{intermediate} + \mathcal{L}_\text{NPCS}$. For more details about the intermediate auxiliary loss and the training strategy, please refer to \cite{liu2022autogpart}.

\section{More Details on Part-based Object Manipulation}

\subsection{Interaction Policy}
\label{appendix:interaction}

(1) Round Fixed Handle: For a round fixed handle, we use the gripper to approach the handle from the positive direction of the $z$ axis, open the gripper to a width that exceeds the side length of the bounding box, and then close the gripper to complete the grasping.

(2) Line Fixed Handle: The interaction policy for a line fixed handle is similar to a round fixed handle. Note that we want the opening direction of our gripper and the line fixed handle to be perpendicular, so we turn the opening direction parallel to the $y$ axis of the predicted bounding box.

(3) Hinge Handle: The interaction policy for a hinge handle is similar to a line fixed handle. After approaching and grasping the hinge handle, we can rotate it along the predicted axis of the revolute joint.

(4) Slider Button: For a slider button, we close the gripper, approach the button from the positive direction of the $z$ axis, and then press the button.

(5) Hinge Knob: For a hinge knob, we clamp the knob like a round fixed handle and rotate the end-effector to complete the manipulation.

(6) Slider Drawer: A gripper approaches an open drawer along the $z$ axis to fetch something in the drawer, and approaches a drawer against the $x$ axis to open it. More often than not, we expect to grab a handle hopefully located on the front face of a drawer.

(7) Hinge Door: For a hinge door with a handle on the front face, we try to grab the handle to open the door. After grabbing the handle, the gripper rotates around the predicted shaft of the door to complete the opening or closing. For a door without any handles, if the door is not closed, we use the gripper to clamp the outer edge along the $y$ axis of the bounding box to open the door.

(8) Hinge Lid: for a hinge lid, we use an interaction policy similar to a hinge door.

(9) Slider Lid: for a slider lid with a handle, we grab the handle to open the lid. Otherwise, we use the gripper to clamp the edge of the lid along the $x$-$y$ plane of the bounding box, and then move up and down along the $z$ axis to open and close the lid.

\subsection{Simulation Experiments}

\textbf{Benchmark Settings.} We set up our interaction environment using the SAPIEN\cite{xiang2020sapien} simulator, modified from the ManiSkill challenge\cite{mu2021maniskill}. We benchmark our method on 4 tasks, \ie, using a single Franka gripper to open a drawer, open a door, manipulate a handle, and press a button. These tasks exemplify robot manipulation under the motion constraint of a prismatic or a revolute joint. For evaluation, we randomly pick unseen objects that contain doors, drawers, handles, and buttons from seen object categories. Considering the limitation of the single gripper, we select such objects that, given the ground truth of their segmentation and pose, can be opened successfully using the heuristics under our benchmark setting. Furthermore, to evaluate the cross-category generalizability of our method, we also randomly pick unseen objects from previously unseen object categories. Compared to the ManiSkill Challenge\cite{mu2021maniskill}, we limit our observation to a first-frame-only partial point cloud of the object, with only one point around the part center indicating which part to interact with. Given the initial state of the robot, it performs the whole manipulation only based on the observation at the first time step. The action space of the robot is the motor command of the 6 joints of the robot to determine the pose of the gripper, and we use position control to open or close grippers. A success in opening the drawer, opening the door, using the handle, and pressing the button is defined as manipulating the part for 90\% of the motion range within 1,000 steps with a stable stop at the end. For each task, we use 20 objects from seen categories and 20 from unseen categories to construct our benchmarks, respectively. Overall, we conduct 4 manipulation tasks in the simulator with 160 objects from 6 seen object categories and 6 unseen object categories.

\textbf{Part-pose-based Manipulation Heuristics.}
We use the interaction policy based on the heuristics introduced in \cref{appendix:interaction} to open drawers, open doors, manipulate handles, and press buttons. Specifically, when we get the part pose, we can immediately get the grasping pose with our policy. Then we use a motion planning library (\ie, mplib, provided by SAPIEN \cite{xiang2020sapien}), to move our gripper to the grasping pose. Then, with our interaction policy and axis predicted from our method, we design the end-effector trajectory just along the trajectory of the part moving and interpolate the trajectory with a time step of $\frac{1}{250}$. With the IK (Inverse Kinematics) algorithm and a PID controller, we solve the poses of joints and move the end-effector along the trajectory. All of our implementations are decoupled from ROS and can be easily implemented in other simulators.

\textbf{Baselines for Object Manipulation.}
(1) Where2act\cite{mo2021where2act} (Oracle input for the first two tasks). We modified the Where2act interaction pipeline to finish our tasks. We use a similar pulling motion for the first three tasks and a pushing motion for the fourth task. Giving only a point to indicate the part to be interacted with makes it challenging for Where2act to perform proper actions, especially for opening drawers and doors. We thus provide additional information (\ie, the handle center of the target door or drawer as a special indicator), and we directly select this point as the point to be interacted with. Then, after motion direction selection, the action is performed to finish the task. We constrain $N_\text{w2a} = 10$ action steps to finish these tasks.
(2)ManiSkill\cite{mu2021maniskill} (Oracle baseline). ManiSkill provides a method for similar vision-based tasks in a reinforcement learning setting. To satisfy the settings in this baseline, we further provide oracle inputs (\ie, per-frame point cloud observations and ground truth part masks). We also design similar dense rewards for each task and train the policy with the same hyper-parameters as ManiSkill. Please refer to \cite{mu2021maniskill} for more details.


\subsection{Real-world Experiments}

\textbf{Implementation Details.}
To evaluate the robustness and generalizability of our method, we use two robot arms (\ie, KINOVA and FRANKA) to manipulate previously unseen objects with only partial point cloud observations in the real world. We use similar motion planning and a similar end-effector trajectory as what we do in the simulator. A partial point cloud of the target object is acquired from the RGB-D camera (Okulo P1 ToF sensor in our experiments). To set up the interaction environment, we place the object and the robot arm in a proper position for interaction and use ArUco markers to calibrate the camera sensor. We also provide a point indicating the part to interact with, just like in the simulator. During manipulation, we first estimate the bounding box of the target part and calculate the trajectory using the heuristics, then use the control API provided by the robot arm to follow the trajectory and finish the task. Overall, we conduct 4 manipulation tasks, \ie, opening doors, opening drawers, lifting lids, and pressing buttons, in the real world with 11 objects from 2 seen object categories and 3 unseen categories.

\begin{table*}[t]
\label{appendix:sapienresult}
\centering
\begin{tabular}{ c|cc|cc|cc|cc }
\hline
\multirow{2}{*}{Success Rate(\%)}&\multicolumn{2}{c|}{Drawer}  & \multicolumn{2}{c|}{Door}  &\multicolumn{2}{c|}{Handle}  & \multicolumn{2}{c}{Button} \\ 
 &Seen & Unseen &Seen & Unseen &Seen & Unseen &Seen & Unseen \\ \hline
\multicolumn{1}{c|}{Where2act\cite{mo2021where2act}}  & 69.9  & 54.5 & 44.4 & 18.2 & 78.7 & 49.2 & 82.2 & 80.9 \\ 
\multicolumn{1}{c|}{ManiSkill\cite{mu2021maniskill}} & 32.9 & 26.6 & 27.8 & 28.3 & 53.9 & 42.1 & 65.5 & 54.5\\
\multicolumn{1}{c|}{Ours}  & \textbf{95.0} & \textbf{90.0} & \textbf{70.0} & \textbf{55.0} &
\textbf{90.0}&\textbf{85.0}&\textbf{100.0}&\textbf{95.0} \\ \hline
\end{tabular}
\caption{\textbf{Results for Cross-category Object Manipulation in SAPIEN Simulator \cite{xiang2020sapien}.}}
\label{table:environment}
\end{table*}

\begin{figure}[h]
    \centering
    \includegraphics[width=0.99\linewidth]{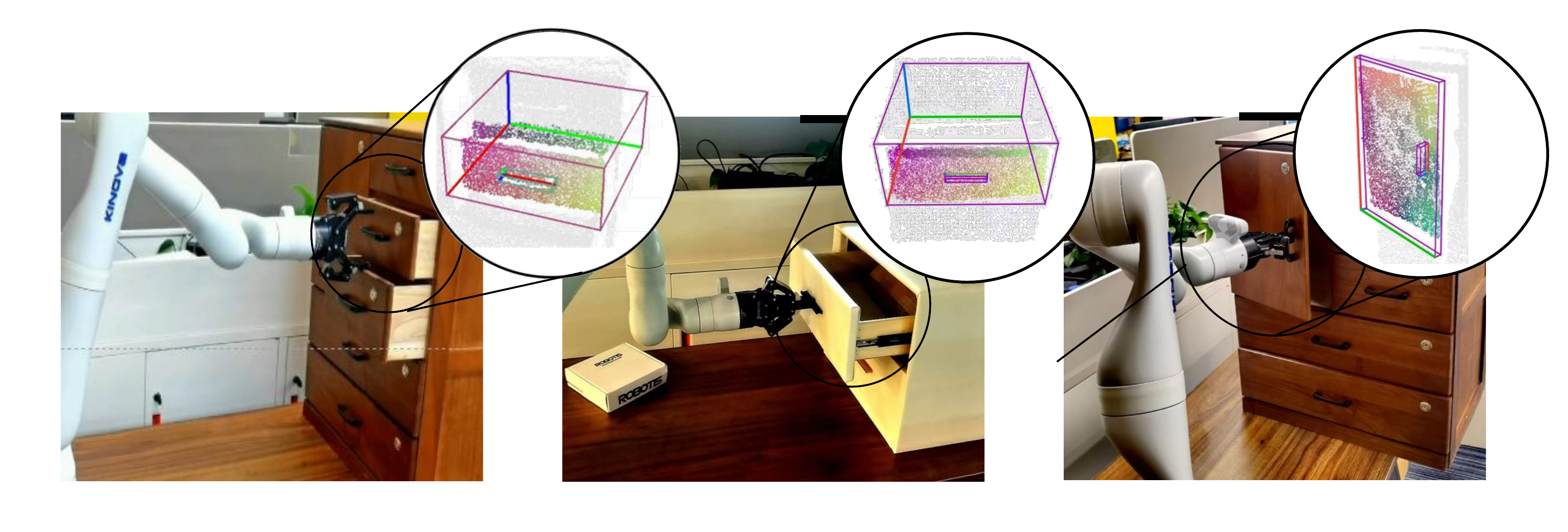}
    \caption{\textbf{More Qualitative Results for Part-based Object Manipulation in the Real-world.}}
    \label{fig:more_real}
\end{figure}

\section{Visualization of GAPartNet Dataset}
\begin{figure*}[ht]
    \centering
    \includegraphics[width=0.9\linewidth]{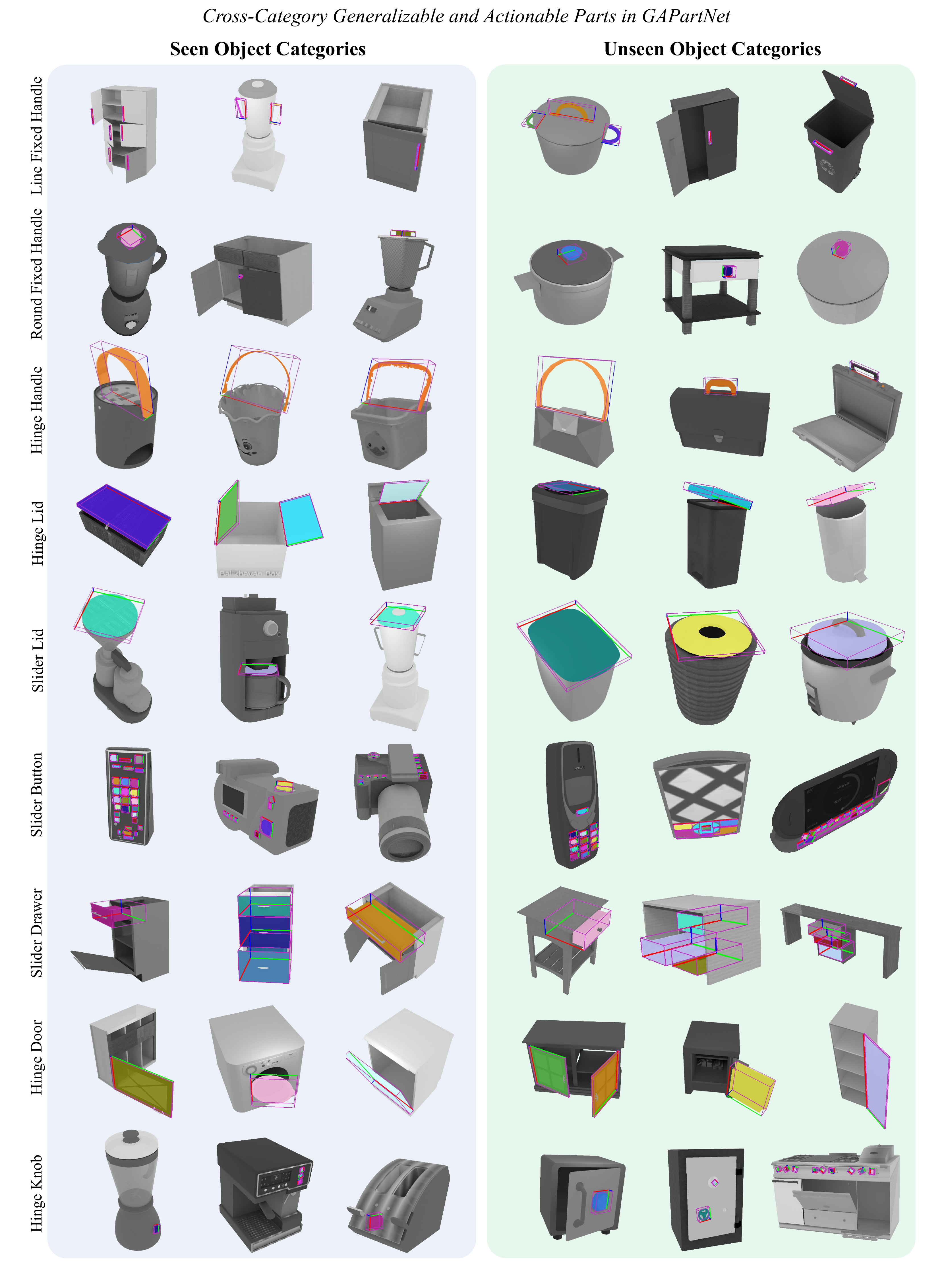}
    \caption{\textbf{Exemplar Objects of Each GAPart Class from Seen Categories and Unseen Categories.} We show objects in gray scale, GAPart segmentation masks in color, and GAPart poses using oriented tight bounding boxes.}
    \label{fig:supp_dataset_gallery}
\end{figure*}

Exemplar objects of each GAPart class from seen categories and unseen categories in the GAPartNet dataset are shown in \cref{fig:supp_dataset_gallery}.

\section{More Results of Part Segmentation and Part Pose Estimation }
\begin{figure*}[t]
    \centering
    \includegraphics[width=1\linewidth]{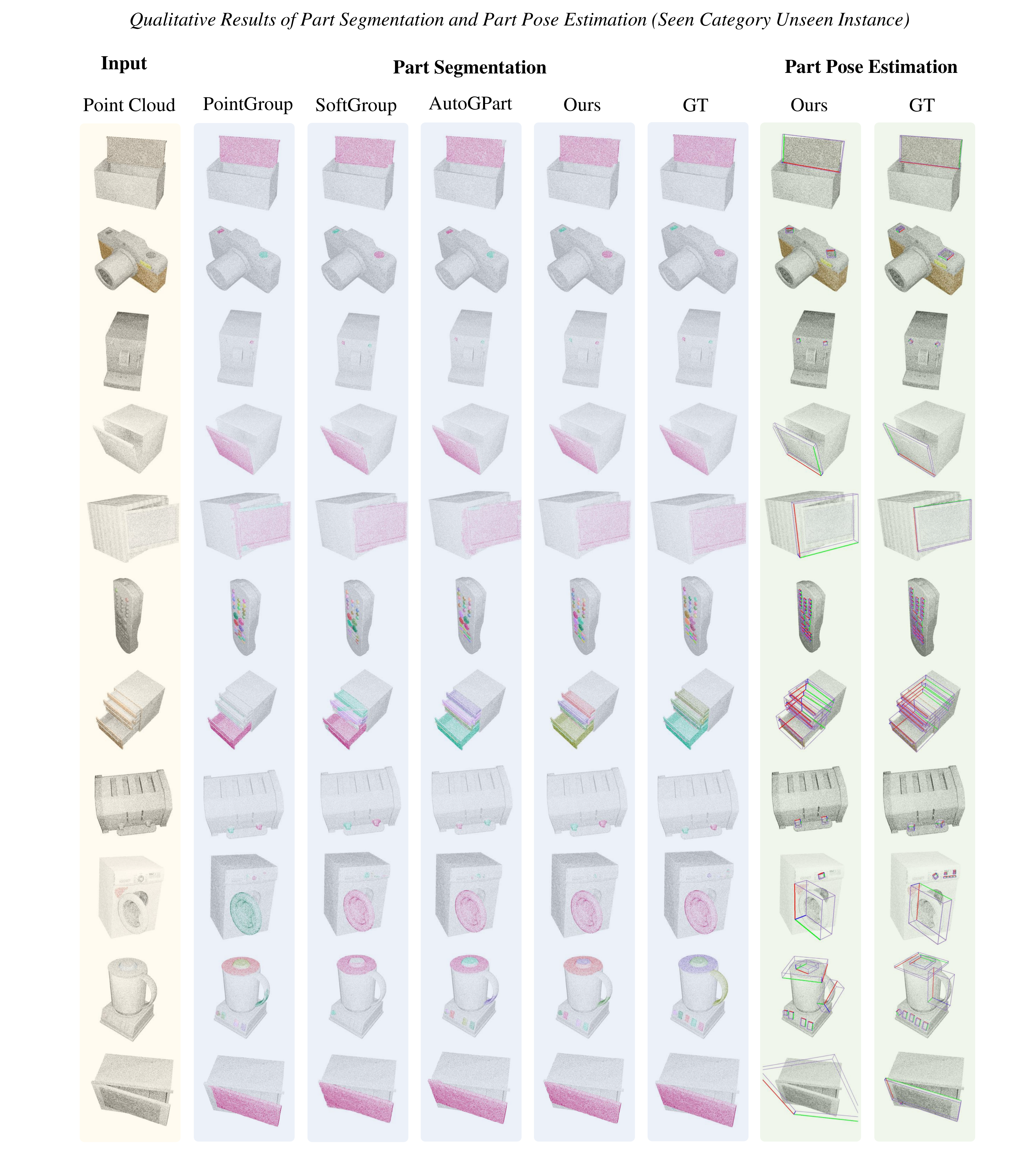}
    \caption{\textbf{Part Instance Segmentation and Pose Estimation Results on the Unseen Instances from the Seen Categories.} Here we compare our method on part instance segmentation task with PointGroup \cite{jiang2020pointgroup}, SoftGroup\cite{vu2022softgroup}, and AutoGPart (modified from \cite{liu2022autogpart}).}
    \label{fig:gallery_result_val}
\end{figure*}

\begin{figure*}[t]
    \centering
    \includegraphics[width=1\linewidth]{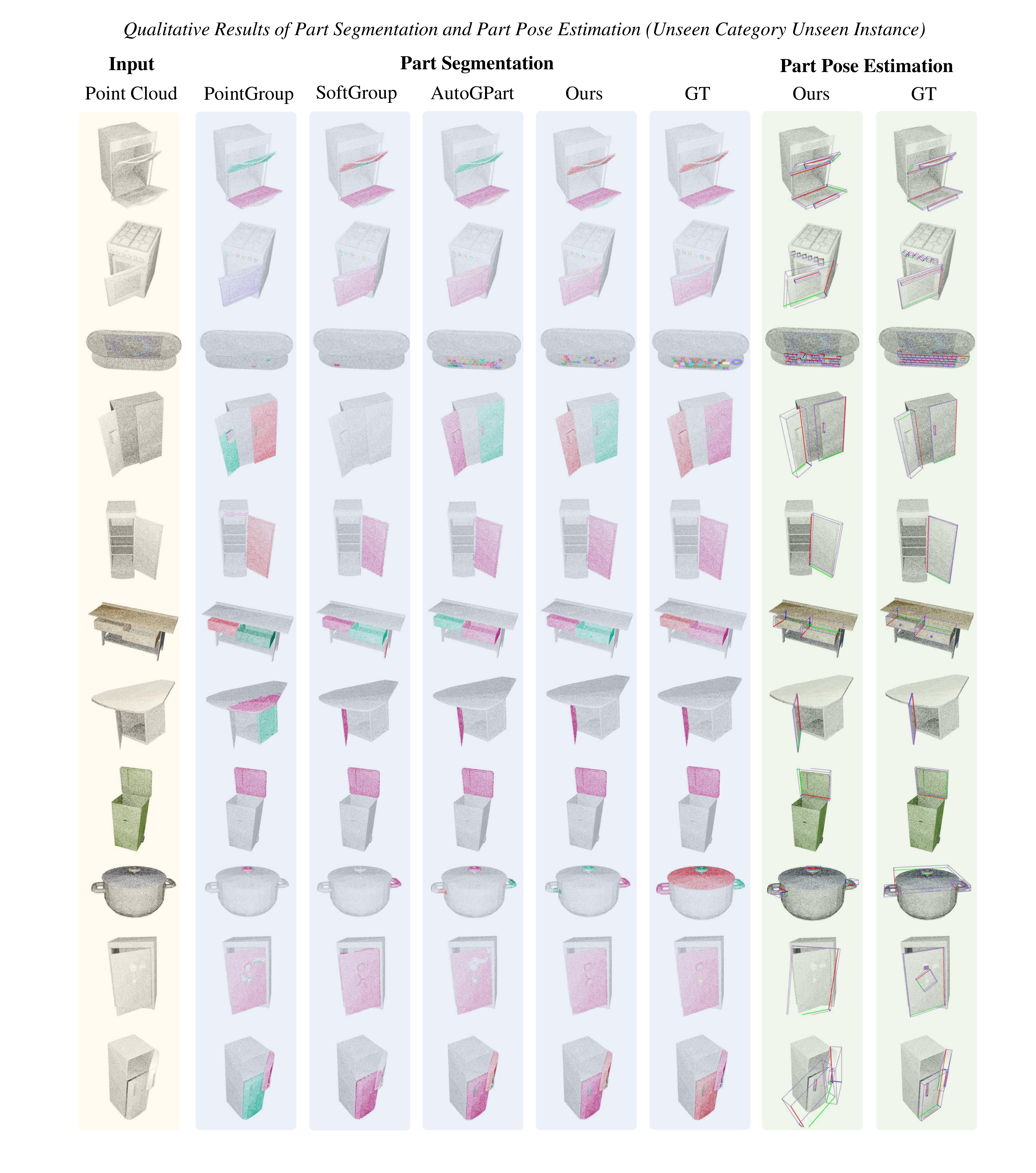}
    \caption{\textbf{Part Instance Segmentation and Pose Estimation Result on the Unseen Instances from the Unseen Categories.} Here we compare our method on part instance segmentation task with PointGroup \cite{jiang2020pointgroup}, SoftGroup\cite{vu2022softgroup}, and AutoGPart (modified from \cite{liu2022autogpart}).}
    \label{fig:gallery_result_test}
\end{figure*}

\begin{figure*}[t]
    \centering
    \includegraphics[width=1\linewidth]{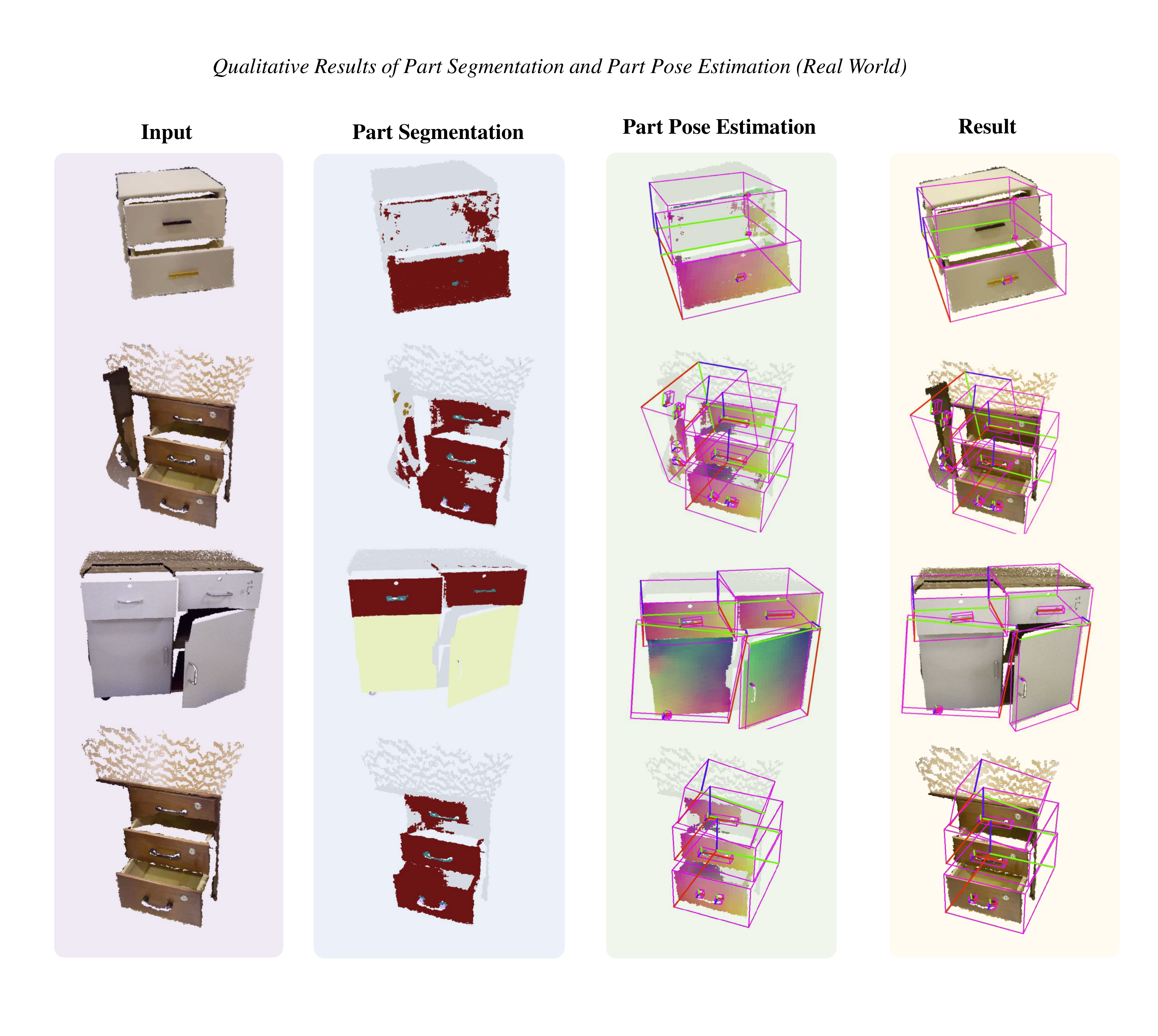}
    \caption{\textbf{Part Instance Segmentation and Pose Estimation Result on the Unseen Objects from the Real World.}}
    \label{fig:gallery_result_real}
\end{figure*}

We visualize more results of part segmentation and part pose estimation in \cref{fig:gallery_result_val,fig:gallery_result_test,fig:gallery_result_real}.

\section{More Results of Part-based Object Manipulation}

For the simulation experiments, the quantitative results are shown in~\cref{table:environment}. Our method significantly outperforms the baselines on all 4 tasks, showing good generalizability and proving the effectiveness of our part-pose-based manipulation policy. More qualitative results are provided in the video 04:30-04:46 on our project page.

For the real-world experiments, more qualitative results are provided in  ~\cref{fig:more_real} and the video 04:47-05:12 on our project page.

\section{More Discussions}

\subsection{Real Depth Signal and Sim-to-Real Gap}
In our experiments, we find that depth quality is crucial to our perception and downstream manipulation. Actually, for our real-world experiments, we have to spray the contrast aid paint onto the transparent lid and use a structural sensor to closely scan the remote and the calculator for obtaining good and detailed geometry.
For diffuse objects with okay depth quality, we argue that further leveraging domain adaptation would be beneficial; however, for certain metallic or transparent objects, their depth will be incomplete, falling into a completely different problem. We leave a more fundamental solution to predict/refine geometry for future works.

\subsection{Outlier Part Shapes}
In our work, GAParts are defined to be functional parts with similar geometry and actionability. So how can our framework tackle the parts with outlier shapes?

Here we take the curvy or irregular handles on doors as an example.
For certain handles, their perception is basically an out-of-distribution perception problem and can theoretically be tackled within our framework; however, we admit the pose of those outliers may not be so informative, which may lead to failure in manipulation heuristics.
We argue that the function and actionability of outlier door handles, \eg, revolving to open, is still the same as the regular ones.
So learning a manipulation policy based on actionable information instead of relying on heuristics would be promising (see our further discussion in \cref{appendix:Pose_For_RL}) and can potentially handle those outliers. 

\subsection{Part Information for Manipulation in RL}
\label{appendix:Pose_For_RL}
By definition, the GAPart carries abundant information about the part's pose, function, actionability, \etc, which is valuable to facilitate manipulation policy learning in RL. 
Here we provide a pilot study and some preliminary results to showcase the usefulness of GAPart information. 
We conduct experiments on learning cross-category manipulation policy from state observations for opening door and opening drawer tasks using PPO under dense rewards, as shown in \cref{table: RL-pose}.
We take proprioceptive information and the bounding box of the door/drawer as state input. The distinction between w/ and w/o pose is whether an additional state input, ground truth handle pose, is used.
The results demonstrate that oracle GAPart information can significantly benefit policy learning. 
This would hopefully shed light on more advanced RL designs in future research, such as incorporating part pose estimation into reward functions and leveraging the part pose to canonicalize visual signals.

\begin{table}[h]
\centering
\resizebox{0.88\linewidth}{!}{
    \begin{tabular}{c|c|ccc}
    \hline  &  &  Train  & \multicolumn{2}{c} {Test Set} \\ 
    \cline{4-5}
    &  & Set &  S.C. &  U.C. \\ 
    \hline Opening &  w/o Pose & 26.1±4.9 &22.0±2.3 &18.1±2.8  \\
       Door  & w/ Pose & \textbf{58.3±3.9} & \textbf{37.9±2.5} & \textbf{18.3±2.9} \\
    \hline Opening &  w/o Pose & 59.8±4.2 & 40.9±4.5 & 18.4±3.3 \\
       Drawer  & w/ Pose & \textbf{91.2±5.2} & \textbf{87.1±6.7} & \textbf{35.6±3.8} \\
     \hline
    \end{tabular}
    }
\caption{\textbf{Part poses improve RL success rate.} S.C.=Unseen objects in seen categories. U.C.=Unseen objects in unseen categories. A larger benchmark for RL with
\# instances: 258/63/77 for doors, 138/57/96 for drawers.
}
\label{table: RL-pose}
\end{table}

\end{document}